\documentclass[10pt,journal,compsoc]{IEEEtran}
\usepackage{graphicx}
\usepackage{amssymb}
\usepackage{subfigure}
\usepackage{xcolor}
\usepackage{url}
\usepackage{multirow}
\ifCLASSOPTIONcompsoc
  \usepackage[nocompress]{cite}
\else
  \usepackage{cite}
\fi
\usepackage{amsmath}
\usepackage{ragged2e}
\ifCLASSINFOpdf
\else
\fi
\begin{document}

\title{A Survey on Large-scale Machine Learning}
\author{Meng Wang,~\IEEEmembership{Senior Member,~IEEE,}
        Weijie Fu,~
		Xiangnan He,~
		Shijie Hao,~
        and~Xindong Wu,~\IEEEmembership{Fellow,~IEEE}
\IEEEcompsocitemizethanks{\IEEEcompsocthanksitem Meng Wang, Weijie Fu, Shijie Hao and Xindong Wu are with the Key Laboratory of Knowledge Engineering with Big Data (Hefei University of Technology), Ministry of Education, and the School of Computer Science and Information Engineering, Hefei University of Technology, Hefei, Anhui, 230601, China. E-mail: \{eric.mengwang, fwj.edu, hfut.hsj\}@gmail.com and xwu@hfut.edu.cn.
\IEEEcompsocthanksitem Xiangnan He is with the University of Science and Technology of China, Hefei, Anhui, 230031, China. E-mail: xiangnanhe@gmail.com.}
\thanks{\qquad }} 
\markboth{Accepted by IEEE Transactions on Knowledge and Data Engineering}
{Shell \MakeLowercase{\textit{et al.}}: Bare Demo of IEEEtran.cls for Computer Society Journals}
\IEEEtitleabstractindextext{
\begin{abstract}
\justifying
Machine learning can provide deep insights into data, allowing machines to make high-quality predictions and having been widely used in real-world applications, such as text mining, visual classification, and recommender systems. However, most sophisticated machine learning approaches suffer from huge time costs when operating on large-scale data. This issue calls for the need of {Large-scale Machine Learning} (LML), which aims to learn patterns from big data with comparable performance efficiently. In this paper, we offer a systematic survey on existing LML methods to provide a blueprint for the future developments of this area. We first divide these LML methods according to the ways of improving the scalability: 1) model simplification on computational complexities, 2) optimization approximation on computational efficiency, and 3) computation parallelism on computational capabilities. Then we categorize the methods in each perspective according to their targeted scenarios and introduce representative methods in line with intrinsic strategies. Lastly, we analyze their limitations and discuss potential directions as well as open issues that are promising to address in the future.
\end{abstract}
\begin{IEEEkeywords}
large-scale machine learning, efficient machine learning, big data analysis, efficiency, survey
\end{IEEEkeywords}}

\maketitle
\IEEEdisplaynontitleabstractindextext
\IEEEpeerreviewmaketitle
\section{Introduction}\label{sec:introduction}
Machine learning endows machines the intelligence to learn patterns from data, eliminating the need for manually discovering and encoding the patterns. Nevertheless, many effective machine learning methods face quadratic time complexities with respect to the number of training instances or model parameters \cite{duda2012pattern}. With the rapidly increasing scale of data in recent years \cite{wu2014data}, these machine learning methods become overwhelmed and difficult to serve for real-world applications. To exploit the gold mines of big data, {Large-scale Machine Learning} (LML) is therefore proposed. It aims to address the general machine learning tasks on available computing resources, with a particular focus on dealing with large-scale data. LML can handle the tasks with nearly linear (or even lower) time complexities while obtaining comparable accuracies. Thus, it has become the core of big data analysis for actionable insights. For example, self-driving cars such as Waymo and Tesla Autopilot apply convolutional networks in computer vision to perceive their surroundings with real-time images \cite{kim2017interpretable}; online media and E-commerce sites such as Netflix
and Amazon build efficient collaborative filtering models from users' histories to make product recommendations \cite{bayer2017generic}. All in all, LML has been playing a vital and indispensable role in our daily lives.

Given the increasing demand for learning from big data, a systematic survey on this area becomes highly scientific and practical. Although some surveys have been published in the area of big data analysis \cite{al2015efficient,bottou2018optimization,tsai2015big,chen2014bigB}, they are less comprehensive in the following aspects. Firstly, most of them only concentrate on one perspective of LML and overlook the complementarity. It limits their values for understanding this area and promoting future developments. For example, \cite{al2015efficient} focuses on predictive models without covering the optimization, \cite{bottou2018optimization} reviews stochastic optimization algorithms while ignoring the parallelization, and \cite{tsai2015big} only pays attention to processing systems for big data and discusses the machine learning methods that the systems support. Secondly, most surveys either lose the insights into their reviewed methods or overlook the latest high-quality literature. For example, \cite{al2015efficient} lacks discussions on the computational complexities of the reviewed models, \cite{bottou2018optimization} neglects the optimization algorithms that address the data with high dimensionality, and \cite{landset2015survey} limits its investigation to distributed data analysis in the Hadoop ecosystem.

In this paper, we thoroughly review over 200 papers on LML from computational perspectives with more in-depth analysis and discuss future research directions. We provide practitioners lookup tables to choose predictive models, optimization algorithms, and processing systems based on their demands and resources. Besides, we offer researchers guidance to develop the next generation of LML more effectively with the insights of current strategies. We summarize the contributions as follows.

Firstly, we present a comprehensive overview of LML according to three computational perspectives. Specifically, it consists of: 1) model simplification, which reduces computational complexities by simplifying predictive models; 2) optimization approximation, which enhances computational efficiency by designing better optimization algorithms; and 3) computation parallelism, which improves computational capabilities by scheduling multiple computing devices.

Secondly, we provide an in-depth analysis of existing LML methods. To this end, we divide the methods in each perspective into finer categories according to targeted scenarios. We analyze their motivations and intrinsic strategies for accelerating the machine learning process. We then introduce the characteristics of representative achievements. In addition, we review the hybrid methods that jointly improve multiple perspectives for synergy effects.

Thirdly, we analyze the limitations of the LML methods in each perspective and present the potential directions based on their extensions. Besides, we discuss some open issues in related areas for the future development of LML.

The paper is organized as follows. We first present a general framework of machine learning in Sec.2, followed by a high-level discussion on its effectiveness and efficiency. In Sec.3, we comprehensively review state-of-the-art LML methods and provide in-depth insights into their benefits and limitations. Lastly, we discuss the future directions to address the limitations and other promising open issues in Sec.4, before concluding the paper in Sec.5.

{\section{From Effectiveness to Efficiency}
In this section, we present three perspectives to quantify the efficiency based on an acknowledged error decomposition on effectiveness. Then we provide a brief of LML.

\subsection{Overview of Machine Learning}
\vspace{0.1cm}
\noindent
\textbf{Notations.}
We consider a general setting of machine learning tasks: given $n$ instances $\mathcal{X}=\{\mathbf{x}_1,\ldots,\mathbf{x}_n\}$ sampled from a $d$-dimension space, and ${\{\mathbf{y}_i\}}^{n_L}_{i=1}$ indicate to the labels of the first  ${n_L}$ instances $({n_L}\ge0)$ within $c$ distinct classes. The goal of machine learning is to learn an instance-to-label mapping model $f: \mathbf{x} \to \mathbf{y}$ from a family of functions $\mathcal{F}$, which can handle both existing and future data.

\begin{figure*}[tbp]
\centering
\includegraphics[scale=0.3]{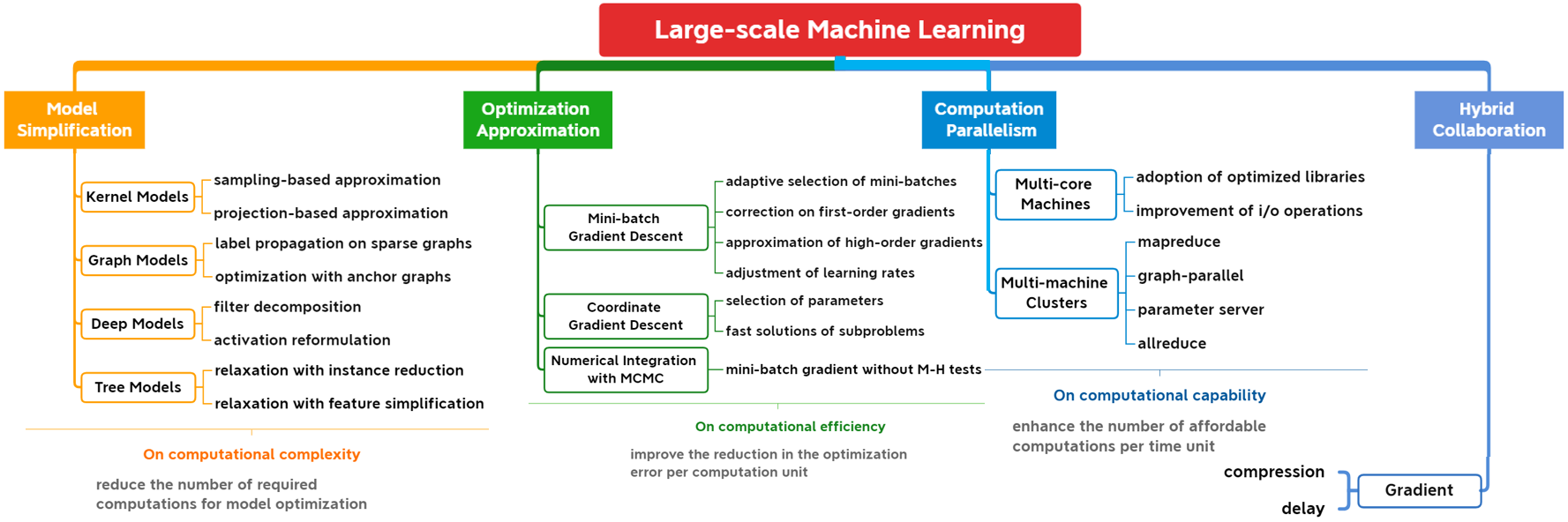}
\caption{\label{Fig:MGD}{The framework of Sec.3, which shows the ways of scaling up machine learning to large-scale machine learning from three perspectives.}}
\label{Fig:categorization}
\end{figure*}

\vspace{0.1cm}
\noindent
\textbf{Effectiveness with error decomposition.}
Let $Q{(f)}_n$ denote the empirical risk of a machine learning model trained over $n$ instances, and $Q{(f)}$ be the corresponding expected risk where the number of instances is infinite. Then we introduce the following specific functions obtained under different settings. Specifically, let $f^*$$=$$\mathrm{argmin}_f Q(f)$ be the optimal function that may not belong to the function family $\mathcal{F}$, and $f^*_{\mathcal{F}}$$=$$\mathrm{argmin}_{f\in\mathcal{F}}Q(f)$ be the optimal solution in $\mathcal{F}$. Suppose $f_n$$=$$\mathrm{argmin}_{f\in\mathcal{F}}Q(f)_n$ is the optimal solution that minimizes the empirical risk $Q{(f)}_n$, and $\tilde{f}_n$ refers to its approximation obtained by iterative optimization.

Let $T(\mathcal{F},n,\rho)$ be the computational time for an expected tolerance $\rho$ with $Q(\tilde{f}_n)_n$$-Q(f_n)_n$$<$$\rho$. Based on the above definitions, the excess error $\mathcal{E}$ obtained within an allotted time cost $T_{max}$ can be decomposed into three terms \cite{bottou2008tradeoffs}:
\begin{eqnarray}\label{eq:ErrorDecomposition}
\mathrm{argmin}_{\mathcal{F},n,\rho} \;\mathcal{E}_{app}+\mathcal{E}_{est}+\mathcal{E}_{opt}, \;
\mathrm{s.t.}  \; T(\mathcal{F},n,\rho)\le T_{max}
\end{eqnarray}
where $\mathcal{E}_{app}$=$\mathbb{E}[Q{(f^*_{\mathcal{F}})}$$-$$Q{(f^*)}]$ denotes the approximation error, measuring how closely the functions in $\mathcal{F}$ can approximate the optimal solution beyond $\mathcal{F}$; $\mathcal{E}_{est}$=$\mathbb{E}[Q{(f_n)}$$-$ $Q{(f^*_{\mathcal{F}})}]$ is the estimation error, which evaluates the effect of minimizing the empirical risk instead of the expected risk; $\mathcal{E}_{opt}$=$\mathbb{E} [Q{(\tilde{f}_n)}$$-$$Q{(f_n)}]$ indicates the optimization error, which measures the impact of the approximate optimization on the generalization performance.

\vspace{0.1cm}
\noindent
\textbf{Efficiency from three perspectives.} Based on a low degree of reduction of the above decomposition, we show the three perspectives for improving machine learning efficiency.

Firstly, we focus on the effect of $\mathcal{E}_{app}$, which is predefined by the function family $\mathcal{F}$ of predictive models. By tuning the size of $\mathcal{F}$, we can make a trade-off between $\mathcal{E}_{app}$ and the computational complexity. Secondly, we consider the influence of $\mathcal{E}_{est}$. According to the probably approximately correct theory, the required number of instances for optimizing models in a large-size function family can be much larger \cite{hanneke2016optimal}. To make use of available data and reduce the estimation error $\mathcal{E}_{est}$,  we suppose all $n$ instances are traversed at least once during optimization. Thus, we ignore $\mathcal{E}_{est}$ and the factor $n$.  Thirdly, we pay attention to $\mathcal{E}_{opt}$,
which is affiliated with optimization algorithms and processing systems. Specifically, the algorithms play a crucial role in computational efficiency, which attempts to increase the reduction in the optimization error per computation unit. The processing systems determine the computational capacity based on the hardware for computing and the software for scheduling. With a powerful system, plenty of iterations in optimization can be performed within allotted time costs.

Above all, the machine learning efficiency can be improved from three perspectives, including 1) the computational complexities of predictive models, 2)  the computational efficiency of optimization algorithms, and 3) the computational capabilities of processing systems.

\subsection{Brief of Large-scale  Machine Learning}
Now we present a brief of LML based on the above analysis.}

1) Reduce the computational complexity based on model simplification.
Machine learning models can be optimized based on matrix operations, which generally take cubic computational complexities for square matrices. Besides, most non-linear methods require extra quadratic complexities for estimating the similarity between pairs of instances to formulate their models. If these models are simplified by being constructed and optimized with smaller or sparser involved matrices, we can reduce their complexities significantly.

2) Improve the computational efficiency based on optimization approximation. Gradient descent algorithms cannot always compensate for their huge computations by selecting larger learning rates \cite{bottou2018optimization}. Therefore, a more effective manner is splitting data or parameters into multiple subsets and then updating models with these small subsets. As these approximate algorithms can obtain relatively reliable gradients with fewer computations, the reduction in the optimization error can be increased per computation unit.

3) Enhance the computational capacity based on computation parallelism. During the procedure of model construction and optimization, a high number of computation-intensive operations can be performed simultaneously, such as the calculations that are repeated on different mini-batches. Built upon parallel processing systems, we can break up these intensive computations and accomplish them on multiple computing devices with a shorter runtime.

\section{Review on Large-scale  Machine Learning}
In this section, we review LML in detail. Specifically, we present the methods in the above three perspectives in Sec.3.1-Sec.3.3, and discuss their collaboration in Sec.3.4. For each part, we categorize the related methods according to targeted scenarios and introduce the methods based on their intrinsic strategies. We also provide experimental evidence to demonstrate the effectiveness of these strategies and summarize their pros and cons. For convenience, Fig.\ref{Fig:categorization} provides a coarse-to-fine overview for the structure of this section.

\begin{table*}[tbp]
\centering
\caption{\label{Tab:simplification}{A brief lookup table for LML methods based on Model Simplification.}}
\begin{tabular}{l l l}
\hline\hline
 \textbf{Categories}       &  \multirow{1}{*}{\textbf{Strategies}}& \multirow{1}{*}{\textbf{Representative Methods}} \\ \hline
 \multirow{2}{*}{Kernel-based Models}       & sampling-based approximation &  uniformly sampling \cite{kumar2012sampling,yang2012nystrom}, incremental sampling \cite{bouneffouf2015sampling,farahat2011novel}. \\  \cline{2-3}
& projection-based approximation &  Gaussian  \cite{martinsson2011randomized}, orthonormal transforms \cite{yang2017randomized}, ramdom features \cite{rahimi2008random,lu2016large}. \\ \cline{1-3}
\multirow{2}{*}{Graph-based Models}  & label propagation on sparse graph & approximate search \cite{kalantidis2014locally, zhang2016discrete,Zhang2013Fast}, division and conquer \cite{Chen2009Fast,wang2012scalable}. \\ \cline{2-3}
& optimization with	anchor graph & single layer \cite{liu2010large, wang2016scalable, zhang2015scaling}, hierarchical layers \cite{fu2017flag,wang2017learning}. \\ \cline{1-3}
 \multirow{2}{*}{Deep Models}  & filter decomposition & on channels \cite{howard2017mobilenets, krizhevsky2012imagenet, sifre2014rigid,szegedy2015going}, on spatial fields \cite{simonyan2014very,szegedy2016rethinking, yu2016multi}. \\ \cline{2-3} &activation reformulation& at hidden layers \cite{nair2010rectified, chen2019addernet, liew2016bounded, maas2013rectifier}, at the output layer \cite{mikolov2013distributed, morin2005hierarchical}.\\	\cline{1-3}
 \multirow{2}{*}{Tree-based Models}  & relaxation with  instance reduction &random sampling \cite{friedman2002stochastic}, sparse-based\cite{chen2016xgboost}, importance-based\cite{ ke2017lightgbm}.\\ \cline{2-3}
 & relaxation with feature simplification & random sampling \cite{breiman2001random}, histogram-based \cite{chen2016xgboost,ben2010streaming}, exclusiveness-based \cite{ke2017lightgbm}. \\
\hline\hline
\end{tabular}
\end{table*}

\subsection{Model Simplification}
Model simplification methods improve machine learning efficiency from the perspective of computational complexities. To reformulate predictive models and lower computational complexities, researchers introduce reliable domain knowledge for a small $\mathcal{E}_{app}$, such as the structures or distributions of instances and the objectives of learning tasks \cite{bengio2013representation}. {According to targeted scenarios, the reviewed scalable predictive models consist of four categories, including kernel-based, graph-based, deep-learning-based, and tree-based. For convenience, a brief overview is provided in Tab.\ref{Tab:simplification}.

\subsubsection{For Kernel-based Models}
Kernel methods play a central role in machine learning and have demonstrated huge success in modeling real-world data with highly complex, nonlinear distributions \cite{murphy2012machine}. The key element of these methods is to project instances into a kernel-induced Hilbert space $\phi(\textbf{x})$, where dot products between instances can be computed equivalently through the kernel evaluation as $K_{ij}$=$<$$\phi(\textbf{x}_i),\phi(\textbf{x}_j)$$>$.

\vspace{0.1cm}
\noindent
\textbf{Motivation.} Given $n$ instances, the computational complexity of  constructing a kernel matrix $\mathbf{K}$$\in$$\mathbb{R}^{n\times n}$ scales as $O(n^2d)$. Supposing all instances are labeled with a label matrix of $\mathbf{Y}$$\in$$\mathbb{R}^{n\times c}$ ($n$=$n_L$), most kernel-based methods can be solved based on matrix inversion as
\begin{equation}\label{eq:kernel}
{(\mathbf{K}+\sigma\mathbf{I})}^{-1}\mathbf{Y},
\end{equation}
which requires a computational complexity of $O(n^3+n^2c)$. Examples include the Gaussian  process, kernel ridge regression, and least-square support vector machine \cite{gestel2002bayesian}.

To alleviate the above computational burdens for a large $n$, a powerful solution is performing low-rank approximation based on SPSD sketching models \cite{gittens2016revisiting} and solving the matrix inversion with Woodbury matrix identity \cite{higham2002accuracy}. Specifically, let $\mathbf{S} \in \mathbb{R}^{n\times m}$ with $n\ge m$ indicate the sketching matrix. Take $\mathbf{C}=\mathbf{K}\mathbf{S}$, and $\mathbf{W}=\mathbf{S}^\mathrm{T}\mathbf{K}\mathbf{S}$. Then $\mathbf{C}\mathbf{W}^{\dagger}\mathbf{C}^\mathrm{T}$ becomes a low-rank approximation to $\mathbf{K}$ with the rank at most $m$. Based on the matrix inversion lemma \cite{higham2002accuracy}, Eq.\ref{eq:kernel} can be rewritten into
$\frac{1}{\sigma}[\mathbf{Y}-\mathbf{C}{({\sigma}\mathbf{W} + \mathbf{C}^\mathrm{T}\mathbf{C})}^{-1}(\mathbf{C}^\mathrm{T}\mathbf{Y})]$, which reduces the computational complexity to $O(m^3$+$nmc)$.

For scaling up kernel-based models without hurting the performance, the choice of $\mathbf{S}$ or the efficient construction of $\mathbf{C}$ becomes crucial. Below we introduce two strategies for low-rank approximation based on sketching matrices.

\vspace{0.1cm}
\noindent
\textbf{Sampling-based approximation.} With this strategy, $\mathbf{S}$ represents a sparse matrix that contains one nonzero in each column. A direct method is sampling columns of kernel matrices at random with replacement \cite{yang2012nystrom,kumar2012sampling}, which is equivalent to the Nystr{\"o}m approximation. By assuming that potential clusters are convex, one can select columns corresponding to kmeans centers with the cost of $O(nmdt)$ \cite{zhang2015scaling}, where $t$ is the number of iterations. To weigh the complexity and the performance, a few incremental sampling methods are proposed. At each iteration, these methods first randomly sample a subset of columns and then pick the column either with the smallest variance of the similarity matrix between the sampled columns and the remaining ones or the lowest sum of squared similarity between selected ones  \cite{zhang2008improved, bouneffouf2015sampling,farahat2011novel}.  In general, the computational complexities of these methods scale as $O(nmp)$, where $p$ is the size of the subset. Besides, the structures of clustered blocks can be exploited to reduce storage costs\cite{Si2017Memory}.

\vspace{0.1cm}
\noindent
\textbf{Projection-based approximation.} Based on this strategy, $\mathbf{S}$ is a dense matrix, which consists of random linear combinations of all columns of kernel matrices  \cite{halko2011finding}. For example, one can introduce Gaussian distributions to build a data-independent random projection \cite{martinsson2011randomized}. Besides, the sketch can be improved with orthonormal columns that span uniformly random subspaces. To do this, we can randomly sample and then rescale the rows of a fixed orthonormal matrix, such as Fourier transform matrices, and Hadamard matrices \cite{yang2017randomized}. These orthonormal sketches can additionally speed up the matrix product to $O(n^2\mathrm{log}m)$, as opposed to $O(n^2m)$ required for the same operation with general dense sketches \cite{yang2017randomized}. On the other hand, some methods construct kernel spaces directly by mapping data non-linearly to new low-dimension subspaces  \cite{rahimi2008random,lu2016large,mutny2018efficient}.

\vspace{0.1cm}
\noindent
\textbf{Discussion based on experimental results.} Benefit from matrix identity lemma on the low-rank approximation, time costs can be significantly reduced. For example, \cite{xu2011efficient} sped up the matrix inversion nearly 100+ times without scarifying the accuracy for COREL images. Below we discuss the choice of the above methods in practical applications.

Firstly, for sampling-based methods, uniform sampling is the fastest solution. Although its performance is generally worse than others with a small $m$, it can be mitigated with the increase of $m$ \cite{kumar2012sampling}. Besides, the effectiveness and efficiency of incremental sampling are weighed based on $q$. However, its time cost could be 50\% lower than kmeans, especially when the selected number of columns was small \cite{farahat2011novel}. Secondly, for projection-based methods, Gaussian matrices lead to comparable or better performance than orthonormal sketches \cite{yang2017randomized, gittens2016revisiting}. Thirdly, projection-based sketches are consistently better than uniformly sampling, while the latter outperforms naive random features \cite{lu2016large, yang2012nystrom}. Finally, since sampling-based methods only need to compute the involved similarity, the computation of a full kernel matrix is avoided. However, if data comes in streaming with varying distributions, these methods are no longer suitable \cite{lu2016large}. In contrast, random features can be solved in the primal form through fast linear solvers, thereby enabling to handle large-scale data with acceptable performance \cite{rahimi2008random, sriperumbudur2015optimal}.

\subsubsection{For Graph-based Models}
Graph-based methods define a graph where the nodes represent instances in the dataset, and the weighted edges reflect their similarity. For classification tasks, their underlying assumption is label smoothness over the graph \cite{murphy2012machine}.

\vspace{0.1cm}
\noindent
\textbf{Motivation.} Given $n$ instances, a graph is first constructed to estimate the similarity between all instances, measured as $W_{ij}$=$\mathrm{RBF}(x_i, x_j)$. Then graph-based models constrain the labels of nearby instances to be similar and the predicted labels towards the ground truths. Let $\mathbf{Y}$ be the label matrix where only $n_L$ rows contain nonzero elements. Consequently, the soft label matrix $\mathbf{F}$ can be solved by
\begin{equation}\label{eq:Implicit01}
  \mathbf{F}={(\mathbf{I}+\alpha\mathbf{L})}^{-1}\mathbf{Y},
\end{equation}
where $\mathbf{L}$=$\mathrm{diag}{(\mathbf{1^\mathrm{T}W})}$-$\mathbf{W}$ denotes the graph Laplacian matrix. The computational cost of graph models comes from two aspects: the graph construction with the cost of $O(n^2d)$ and the matrix inversion with the cost of $O(n^3)$. Thus, there are two types of strategies for improving scalability, including label propagation on sparse graphs and optimization with anchor graphs.

\vspace{0.1cm}
\noindent
\textbf{Label propagation on sparse graphs.} Unlike kernels, most graphs prefer the sparse similarity, which has much less spurious connections between dissimilar points \cite{liu2010large}. Thus, we can conduct iterative label propagation to accelerate the spread of labels, represented as
$\mathbf{F}^{t+1}$=$\alpha(\mathbf{W})\mathbf{F}^t$+$(1$-$\alpha)\mathbf{Y}$,
where $\mathbf{F}^0$=$\mathbf{Y}$. Let $k$ denote the average nonzero element in each row of $\mathbf{W}$. Then the number of necessary computations in each iteration scales as $O(nkC)$, taking up a tiny part of the original amount $O(n^2C)$. As a result, a huge part of the computational complexity now comes from graph construction, which can be efficiently solved based on approximate sparse graph construction.

These graph construction methods introduce a hierarchical division on datasets to find the neighbors of each instance and estimate their similarity, which reduces the cost to $O(n\mathrm{log}(n)d)$. For example, approximate nearest neighbor search (ANNS) first builds a structured index with all instances \cite{kalantidis2014locally} and then searches the approximate neighbors of each instance on the obtained index, such as hierarchical trees \cite{wang2017learning} and hashing tables \cite{wang2018survey,zhang2016discrete}. To enhance the quality, we can repeat the above procedure to generate multiple basic graphs and combine them to yield a high-quality one \cite{Zhang2013Fast}. Besides, as sparse graph construction is much easier than the nearest neighbor search, divide-and-conquer methods become more popular. These methods first divide all instances into two or three overlapped subsets and then unite the sub-graphs constructed from these subsets multiple times with neighbor propagation \cite{Chen2009Fast,wang2012scalable}.

\vspace{0.1cm}
\noindent
\textbf{Optimization with anchor graphs.} Anchor graphs sample $m$ instances as anchors and measure the similarity between all instances and these anchors as $\mathbf{Z}\in\mathbb{R}^{n\times m}$. Then the labels of instances are inferred from these anchors, leading to a small set of to-be-optimize parameters. They use anchors as transition nodes to build the similarity between instances for label smoothness, and their soft label matrix can be obtained as $\mathbf{F}$=$\mathbf{Z} (\mathbf{Z}^\mathrm{T}\mathbf{Z}$+$\alpha\tilde{\mathbf{L}})^{-1}\mathbf{Z}^\mathrm{T}\mathbf{Y}$,
where $\tilde{\mathbf{L}}$$\in$$\mathbb{R}^{m\times m} $ is a reduced Laplacian over anchors. Clearly, the cost of graph construction is reduced to $O(nmd)$ or even $O(nd\mathrm{log}(m))$ with ANNS, and the cost of optimization scales as $O(nm^2+m^3)$.

The original anchor graph models introduce sparse adjacency between instances and anchors \cite{liu2010large}. After that, hierarchical anchor graphs propose to retain sparse similarities over all instances while keeping a small number of anchors for label inference \cite{wang2017learning}. In case that the smallest set of anchors still needs to be large and brings considerable computations, FLAG developes label optimizers for further acceleration \cite{fu2017flag}. Besides, EAGR proposes to perform label smoothness over anchors with pruned adjacency \cite{wang2016scalable}.

\vspace{0.1cm}
\noindent
\textbf{Discussion based on experimental results.} Now we show the advantages of the two classes of methods. On the one hand, label propagation is more efficient than the original matrix inversion. Without accuracy reduction, it could obtain 10$\times$ to 100$\times$ acceleration over the matrix inversion \cite{fujiwara2014efficient}. As for approximate graph construction, although hierarchical trees and hashing tables are easy to be used for ANNS \cite{wang2014hashing}, they generally ignore the fact that each query must be one of the instances in the graph construction. Thus, they put redundancy efforts on giving a good result for unnecessary future queries. In contrast, divide-and-conquer methods can obtain higher consistency to the exact graphs with less time cost, which was demonstrated on both Caltech101, Imagenet, and TinyImage \cite{Chen2009Fast}. On the other hand, although the accuracy of anchor graphs is lightly worse than sparse graphs \cite{liu2010large}, they are more potent  on handling very large datasets. The reason is that, once a set of anchors can be stored in the memory, anchor graphs can be efficiently constructed with the memory cost of $O(md+nk)$ rather than $O(nd)$. For example, with hierarchical anchor graphs, the classification on 8 million instances could be implemented on a personal computer within 2 mins \cite{wang2017learning, fu2017flag}.

\subsubsection{For Deep Models}
Deep models introduce layered architectures for data representation \cite{deng2014tutorial}. Instead of fully-connected networks (FCNs) where any pair of input features in each grid-like data are relevant, convolutional neural networks (CNNs) and recurrent neural networks (RNNs) utilize the structures of data by using small-size filters on local receptive fields \cite{krizhevsky2012imagenet}.

\vspace{0.1cm}
\noindent
\textbf{Motivation.}  Take CNN as an example, where each layer of convolution kernels can be viewed as a 4D tensor. Consider input feature maps with the spatial size of $d_I\times d_I$ and the channel number of $m_I$. A standard convolutional layer on the input can be parameterized by convolution kernels with the size of $d_K\times d_K \times m_I \times m_O$, where $d_K$ is the spatial dimension of the square kernel and $m_O$ is the number of output channels. Suppose the stride equals 1. The computational complexity of this convolutional layer becomes
\begin{eqnarray}\label{eq:Implicit01}
d_I\times d_I \times d_K\times d_K \times m_I \times m_O.
\end{eqnarray}
Generally, deep models involve a large number of hidden layers. Although the sizes of filters are much smaller than the sizes of input features, CNNs still contain millions of parameters \cite{krizhevsky2012imagenet}, and the convolutions contribute to the bulk of most computations. Besides, deep models contain two types of bounded activation functions, including those used in hidden layers for feature extraction, and the ones used at output layers for probability prediction. The former type is used for all neurons, and the latter type involves normalization operations. When handling big models or a large number of output neurons, e.g., NLP tasks, both can lead to huge computations during back-propagation \cite{memisevic2010gated}.

To address these issues, we introduce the filter decomposition for simplifying convolution layers and improving the overall speedup. After that, we review efficient activation functions where gradients can be easily estimated.

\vspace{0.1cm}
\noindent
\textbf{Filter decomposition.}
Filter decomposition is derived by the intuition that there is a significant amount of redundancy in 4D tensors. Thus, some methods reduce computations by gathering information from different channels hierarchically. For example, Alexnet develops group convolutions to avoid the computations between two groups of channels \cite{krizhevsky2012imagenet}. Mobilenet introduces separable depthwise convolutions \cite{howard2017mobilenets}, which factorize the filters into purely spatial convolutions followed by a pointwise convolution along with the depth variable \cite{sifre2014rigid}, leading to 70\% reduction of parameters. Inception v1 in turn first implements the dimensionality reduction on the channels of input features \cite{szegedy2015going}, followed by the filters with the original receptive fields. After that, Shufflenet generalizes group convolutions and depthwise convolutions based on the channel shuffle to further reduce the redundancy \cite{zhang2018shufflenet}.

One the other hand, some methods reduce the computations based on the hierarchical information integration over the spatial side. For example, VGG replaces one layer of large-size filters with the two layers of smaller-size filters \cite{simonyan2014very}, reducing nearly 28\% computations. Besides, Inception v3 introduces asymmetric convolutions that decompose the convolution with 3$\times$3 filters into two cascaded ones with the spatial dimensions of 1$\times$3 and 3$\times$1 \cite{szegedy2016rethinking}, and nearly 33$\%$ parameters can be saved. Dilated convolutions further support the exponential expansion of receptive fields without the loss of resolution \cite{yu2016multi}. It shows that a 7$\times$7 receptive field could be explored by two dilated convolutions with 3$\times$3 parameters, reducing 80\% computations.

\vspace{0.1cm}
\noindent
\textbf{Activation reformulation.} The bounded activation functions such as sigmoid and tanh bring expensive exponential operations at each neuron. Besides, both of them face vanishing gradient problems. To address these issues, ReLU releases the bound by simply picking the outputs as max(0, $x_\mathrm{input}$) \cite{nair2010rectified}. Since too many activations being below zero will make ReLU neurons inactive, leaklyReLU additionally introduces small gradients over the negative domain \cite{maas2013rectifier}. Besides, to improve the training stability by constraining the outputs of activations, a few hard bounded functions upon ReLU were developed, such as bounded ReLU and bounded leakyReLU \cite{liew2016bounded}. Moreover, \cite{chen2019addernet} proposes to replace all the multiplications with adds to reduce the computations.

For probability prediction, the implementation of softmax needs to compute a normalization factor based on all output neurons that could be in millions or billions. To reduce the computations, hierarchical softmax reorganizes output probabilities based on a binary tree. Specifically, each parent node divides the own probability to its children nodes, and each leaf corresponds to a probabilistic output  \cite{morin2005hierarchical}. 
When an input-output pair is available, one can only maximize the probability of the path in the binary tree, which reduces the computational complexity logarithmically. When Huffman coding is adopted for an optimal hierarchy, the speedup can be further enhanced \cite{mikolov2013distributed}.

\vspace{0.1cm}
\noindent
\textbf{Discussion based on experimental results.} Gradient descent is widely used for training various networks. Thus, filter decomposition successfully reduces the number of computations by lowering the scale of parameters. For example, Inception v1 was able to become 2$\times$ to 3$\times$ faster than those without dimensionality reduction on channels \cite{szegedy2015going}. Mobilenet with depthwise convolutions could build a 32$\times$ smaller and 27$\times$ less compute-intensive networks than VGG16, which still obtained comparable accuracies \cite{howard2017mobilenets}. Benefit from spatial decomposition, Inception v3 used at least 5$\times$ fewer parameters and achieved 6$\times$ cheaper computationally \cite{szegedy2016rethinking}. Meanwhile, by improving the activations, the computations can be significantly reduced. For example, by using hierarchical softmax on vocabulary with 10,000 words, \cite{morin2005hierarchical} sped up network training more than 250$\times$. If there was no overhead and no constant term, the speedup could be 750$\times$. Besides, the network with ReLU could reach a 25\% training error rate on CIFAR-10 while being 6$\times$ faster than the one with tanh \cite{krizhevsky2012imagenet}.

Above all, it is necessary to combine the above strategies to reduce the computations in iterative optimization. However, to maintain the effectiveness with filter decomposition, it is essential to analyze the complexity of each operation carefully and modify a model progressively \cite{he2015convolutional,chen2018neural}. We also note that, model compression also plays a vital role in deep learning. More details can be found in \cite{cheng2017survey, zhang2018survey}.

{\color{black}\subsubsection{For Tree-based Models}
Tree-based models build hierarchical trees from a root to leaves by recursively splitting the instances at each node using different decision rules \cite{safavian1991survey}, such as Gini index and entropy. After that, random forest (RF) and gradient boosting decision trees (GBDT) introduce ensemble learning to improve the robustness of classifiers and enhance their accuracies. The former trains each tree independently. The latter learns trees sequentially by correcting errors, and often applies second-order approximation for custom losses \cite{friedman2000additive}.

\vspace{0.1cm}
\noindent
\textbf{Motivation.} Given $n$ instances with $d$ features, finding the best split for each node generally leads to the computational complexity of $O(nd)$ by going through all features of each instance. Thus, the datasets with millions of instances and features will lead to huge computational burdens when growing trees. To address this issue, the relaxation of decision rules becomes significantly important, which can be performed from the views of instances and features.

\vspace{0.1cm}
\noindent
\textbf{Relaxation with instance reduction.} Implementing instances sampling while growing trees is the simplest method of relaxing the rules and reducing costs \cite{deng2011fast}, which benefits both trees, RF, and GBDT. Besides, one can take advantage of sparse features to ignore invalid instances when evaluating each split, making the complexity linear to the number of non-missing instances \cite{chen2016xgboost}. To improve the representation of sampled instances for GBDT, GOSS prefers the instances with large gradients \cite{ke2017lightgbm}.

\vspace{0.1cm}
\noindent
\textbf{Relaxation with feature simplification.} Feature sampling is an alternative for rule relaxation, which considers only a subset of features at each split node \cite{breiman2001random}. Besides, histogram-based boosting groups features into a few bins using quantile sketches in either a global or local manner and then perform the splitting based on these bins directly \cite{chen2016xgboost,ben2010streaming}. Since a part of features rarely take nonzero values simultaneously, EFB develops a greedy bundling method to locate these features and then merges them by only extending the range of exclusive features \cite{ke2017lightgbm}.

\vspace{0.1cm}
\noindent
\textbf{Discussion based on experimental results.} Below we discuss the strengths and weaknesses of the relaxation of decision rules. Firstly,
randomly sampling 30\% to 20\% instances could speed up GBDT 3$\times$ to 5$\times$ while improving its performance \cite{friedman2002stochastic}. Sparsity-aware splitting also ran 50 $\times$ faster than the naive version on Allstate-10K \cite{chen2016xgboost}. On the other hand, local histogram aggregation resulted in smaller numbers of iterations than the global one by refining the histogram strategy \cite{chen2016xgboost}. Meanwhile, EFB additionally merged implicitly exclusive features into much fewer features, leading to better performance than sparsity-aware splitting  \cite{ke2017lightgbm}. Secondly, although the relaxed rules significantly speed up the training of tree-based models, it may result in extra costs. For example, GOSS requires the cost of computing gradients, and sparsity-aware splitting has to maintain a nonzero data table with additional memory costs.}

\subsubsection{Summary}
We have reviewed various LML methods from the perspective of computational complexities. Now we discuss both the advantages and disadvantages of the above methods.

Firstly, kernel and graph-based models can be scaled up and optimized more efficiently than deep models. Besides, experts can introduce their domain knowledge on input features and develop specific similarities for these two models. Since the sum of positive semidefinite matrices is still positive semidefinite, it is also easier to merge the similarities of different types of features into a single model \cite{xu2013survey}.  Of note, although graphs generally lead to smaller memory costs than kernels, they are only able to handle the data that is satisfied with the cluster assumption \cite{zhang2015scaling}.

\begin{table*}[tbp]
\centering
\caption{\label{Tab:optimization}A brief lookup table for LML methods based on Optimization Approximation.}
\linespread{5}
\begin{tabular}{lll}
\hline\hline
 \textbf{Categories}       &  \multirow{1}{*}{\textbf{Strategies}}& \multirow{1}{*}{\textbf{Representative Methods}} \\ \hline
  & adaptive sampling of mini-batches  & dynamic batch sizes \cite{smith2017don}, proportional instance sampling \cite{alain2015variance}, \cite{gopal2016adaptive}.  \\ \cline{2-3}
 \multirow{1}{*}{Mini-batch} &correction of first-order gradients  & momentum \cite{qian1999momentum,nesterov2013gradient}, weighted historical gradients,  \cite{schmidt2017minimizing,johnson2013accelerating,defazio2014saga}. \\ \cline{2-3}
\multirow{1}{*}{Gradient Descent}  & approximation of higher-order gradients & mini-batch Quasi-Newton \cite{byrd2016stochastic}, mini-batch Gauss-Newton  \cite{engel2014lsd}. \\\cline{2-3}
&adjustment of learning rates&
refer to iterations \cite{xu2011towards}, to gradients \cite{zeiler2012adadelta,duchi2011adaptive}, to moments  \cite{Kingma2014Adam,reddi2019convergence}.  \\ \cline{1-3}
 \multirow{1}{*}{Coordinate  } & selection of parameters    & Gauss-Seidel manner  \cite{hsieh2008dual,nesterov2012efficiency,chang2008coordinate}, Gauss-Southwel rules \cite{shi2016primer,nutini2015coordinate,dhillon2011nearest}. \\ \cline{2-3}
\multirow{1}{*}{Gradient Descent}& fast solutions of subproblems & using extrapolated points \cite{nesterov2012efficiency,lee2013efficient,lin2014accelerated,schmidt2011convergence,beck2009fast}, caching  \cite{shi2016primer,boyd2011distributed}. \\\hline		
 \multirow{1}{*}{Numerical Integration}  & mini-batch gradient without M-H tests& 1st-order Langevin dynamics \cite{welling2011bayesian,patterson2013stochastic}, 2nd-order Langevin dynamics \cite{chen2014stochastic}. \\ \cline{2-3}
\hline						
\hline	
\end{tabular}
\end{table*}

Secondly, benefit from the hierarchical feature extraction on structural instances, deep models can obtain much higher classification accuracies \cite{russakovsky2015imagenet, baveye2015deep}. However, these methods in turn require huge time costs for training the over-parameterized models. Although filter decomposition methods reduce the computations remarkably, the architectures demand careful designs \cite{szegedy2016rethinking}. Besides, some deep models are mathematically equivalent to kernelized ridge regressions that learn their own kernels from the data \cite{rudin2019stop}. However, they can only build kernels in finite-dimension spaces, and their representations are generally uninterpretable, making predictions hard to be explained \cite{belkin2018understand}. \color{black}{In contrast, tree-based models with hierarchical splitting are more interpretable. Besides, these models can be directly integrated into many other methods for acceleration, such as label trees with binary classifiers \cite{bengio2010label}.}}

\subsection{Optimization Approximation}
Optimization approximation scales up machine learning from the perspective of computational efficiency. In each iteration, these methods only compute the gradients over a few instances or parameters to avoid most useless computations \cite{yuan2012recent}. As a result, they increase the reduction in the optimization error per computation unit and obtain an approximate solution with fewer computations. For a small $\mathcal{E}_{opt}$, advanced mathematical techniques must be used to guarantee the effectiveness of approximation. According to targeted scenarios, we further categorize them into mini-batch gradient descent, coordinate gradient descent, and numerical integration based on Markov chain Monte Carlo. For convenience, an overview is provided in Tab.\ref{Tab:optimization}.

\subsubsection{For Mini-batch Gradient Descent}
The methods of mini-batch gradient descent (MGD)\footnote{MGD in this paper is equal to mini-batch SGD in other papers.} aim to solve the problems with a modest number of parameters but a large number of instances. Compared with stochastic gradient descent, MGD utilizes better gradients estimated over more instances per iteration and generally obtains fast local convergence with lower variances.

\vspace{0.1cm}
\noindent
\textbf{Motivation.} We first review the basic formulation of MGD. Suppose $\mathcal{X}_t$ refers to a mini-batch of instances with the size of $m_t$, and $Q$ denotes the objective function built upon the parameter matrix $\mathbf{W}$. Let $\partial Q(\mathbf{W};\mathbf{x}_{i})$ be the stochastic gradient on $\mathbf{x}_{i}$ and ${\mathbf{G}}_t=\frac{1}{m_t}\sum_{i\in\mathcal{X}_t}\partial Q(\mathbf{W}^{t};\mathbf{x}_{i})$ indicates the aggregated stochastic gradient on $\mathcal{X}_{t}$. Similar to gradient descent, the parameter  $\mathbf{W}$ can be updated by
\begin{eqnarray}
\mathbf{W}^{t+1}=\mathbf{W}^{t}-\eta {\mathbf{G}}_t
\end{eqnarray}
with a learning rate of $\eta$. For large-scale datasets, updating parameters based a few instances leads to large variances of gradients and makes optimization unstable. Although we can estimate gradients with large and fixed batch sizes, it remarkably increases per-iteration costs \cite{bottou2018optimization}.

To address this issue, many LML methods have been proposed by improving  gradient information in each iteration with a few extra computations. According to the roles in MGD, we review the methods from four complementary aspects. An illustrative diagram is shown in Fig.\ref{Fig:MGD}.

\begin{figure}[tbp]
\centering
\includegraphics[scale=0.42]{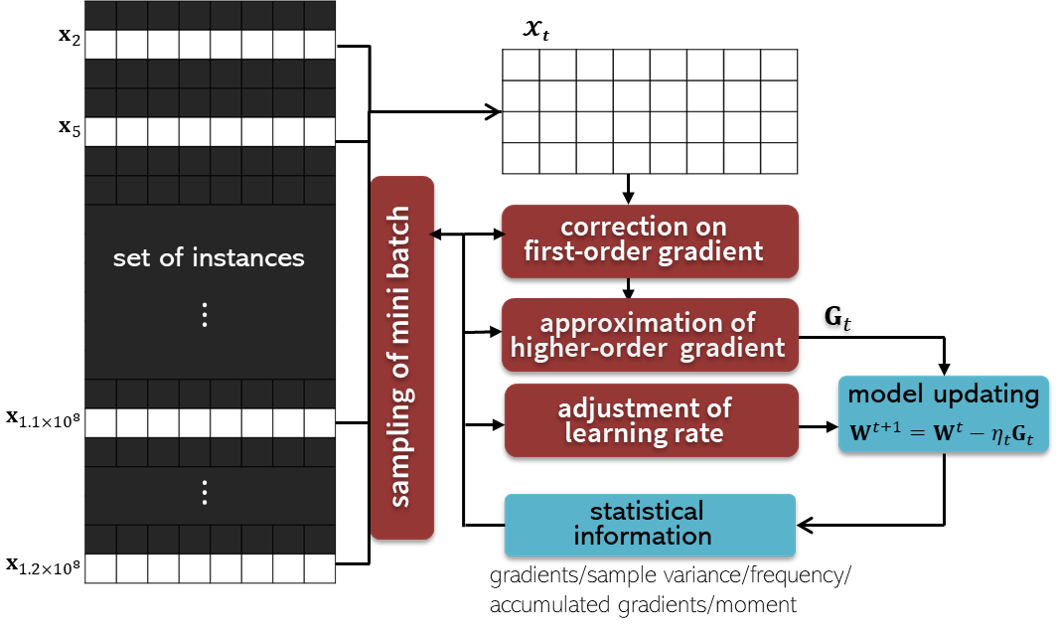}
\caption{\label{Fig:MGD}The diagram of mini-batch gradient descent, where the steps in red are key points for improving the computational efficiency.}
\end{figure}

\vspace{0.1cm}
\noindent
\textbf{Adaptive sampling of mini-batches.} We take account of both the sizes of mini-batches and the sampling of instances. Firstly, since the naive algorithms that partially employ gradient information sacrifice local convergence, we can increase batch sizes gradually via a prescribed sequence \cite{smith2017don}. Besides, linear scaling is effective for large mini-batches \cite{goyal2017accurate}. Secondly, as randomly selected instances are independent of optimization, it is worth considering both data distributions and gradient contributions. Specifically, we can enforce the sampling weights of instances to be proportional to the L2 norm of their gradients \cite{alain2015variance}. Besides, by maintaining a distribution over bins and learning the distribution per $t$ iterations, computations can be further reduced \cite{gopal2016adaptive}.

\vspace{0.1cm}
\noindent
\textbf{Correction of first-order gradients.} Performing the correction on mini-batch gradients provides an alternative to improve the quality of search directions with lower variances, which enables a larger learning rate for accelerations. On the one hand, gradient descent with momentum stores the latest gradients and conducts the next update based on a linear combination of the gradient and previous updates \cite{qian1999momentum}. Gradients descent with Nesterov momentum first performs a simple step towards the direction of the previous gradient and then estimates the gradient based on this lookahead position \cite{nesterov2013gradient}. On the other hand, SAG utilizes the average of its gradients over time to reduce the variances of current gradients \cite{schmidt2017minimizing}, and SVRG develops a memory-efficient version which only needs to reserve the scalars to constrict the gradients at subsequent iterations \cite{johnson2013accelerating}.

\vspace{0.1cm}
\noindent
\textbf{Approximation of higher-order gradients.} When the condition numbers of objective functions become larger, the optimization can be extremely hard owing to ill-conditioning \cite{dauphin2015equilibrated}.  To solve this issue, MGD methods thus introduce the approximation of second-order information with successive re-scaling \cite{ma2017diving,le2011optimization,sun2019survey}. Similar to inexact Newton algorithms, a basic solution is to employ conjugate gradient algorithms to estimate Hessian matrices. However, as the mini-batch size is much smaller, these algorithms implemented in a nearly stochastic manner can lead to very noisy results. To this end, mini-batch-based Quasi-Newton (MQN)  and Gauss-Newton algorithms (MGN) are proposed, which improve the approximation by only using first-order information. In particular, MQN introduces online L-BFGS to approximate the inverse of Hessian based on the latest parameters and a few gradients at previous mini-batches \cite{byrd2016stochastic}. MGN builds the Hessian approximation based on the Jacobian of the predictive function inside quadratic objective functions \cite{engel2014lsd}. When logarithmic losses are used in probability estimation, it enables faster implementation without the explicit Jacobian matrices. It is worthwhile noting that we usually can offset the costs of these approximations when the size of mini-batches is not too small \cite{bottou2018optimization}.

\vspace{0.1cm}
\noindent
\textbf{Adjustment of learning rates.} The learning rate plays an important role in the convergence \cite{ruder2016overview}. Specifically, a small rate may slow down the convergence, while a large rate can hinder convergence and cause the objective function to fluctuate around its minimum. Although the rate decayed by the number of iterations can alleviate this issue \cite{xu2011towards}, recent studies propose to adjust it more carefully according to optimization processes. For example, Adagrad prefers smaller rates for the parameters associated with frequently occurring dimensions and larger rates for the ones associated with infrequent dimensions \cite{duchi2011adaptive}. Besides, Adadelta takes account of the decaying average over past squared gradients \cite{zeiler2012adadelta}. The motivation is that, by restricting the length of accumulated gradients to some fixed sizes, we can reduce the aggressive when decreasing learning rates \cite{bengio2012practical}. Meanwhile, learning rates can also be updated based on the second moment. For example, Adam prefers flat minima in error surfaces  \cite{Kingma2014Adam}. Besides, Nadam and AMSGrad introduce Nesterov-accelerated gradients and the maximum of past squared gradients into Adam, respectively \cite{reddi2019convergence,dozat2016incorporating}.

\vspace{0.1cm}
\noindent
\textbf{Discussion based on experimental results.} Now we discuss the effectiveness of the above strategies. Firstly, adaptive sampling of mini-batch can clearly improve computational efficiency. By doubling batch sizes during training,  \cite{smith2017don} reduced the time cost from 45 mins to 30 mins on ImageNet. Besides, \cite{gopal2016adaptive} introduced adaptive sampling and only used 30\% epochs to obtain the same accuracy of uniform sampling. Since updating the statistic information does not need to be frequent, its extra computational cost is insignificant. Secondly, by correcting gradients for lower variance, SVRG converged faster than the MGD using learning rate scheduling \cite{johnson2013accelerating}. Thirdly, by approximating the second-order information, \cite{ma2017diving} achieved 6$\times$ to 35$\times$ faster when solving eigensystems. MQN also reached a lower objective value with competitive computing time to the original MGD \cite{byrd2016stochastic}. Finally, by taking the frequency of parameters and the decaying of gradients into account for learning rates, Adagrad and Adadelta could only use 20\% epochs to achieve the same result of normal optimizers \cite{zeiler2012adadelta}. Based on adaptive moment estimation, Adam further reduced more than 50\% computations \cite{Kingma2014Adam}. A comprehensive comparison of learning rate schemes can be found in \cite{ketkar2017deep}.  Of note, since these strategies play different roles in optimization, one can apply all for further acceleration \cite{gazagnadou2019optimal}.

\subsubsection{For Coordinate Gradient Descent}
The LML methods built upon coordinate gradient descent (CGD) aim at addressing the problems with a modest number of instances in a high dimension. These problems frequently arise in areas like natural language processing \cite{yuan2012recent} and recommender systems \cite{he2018fast,bayer2017generic}.

\begin{figure}[tbp]
\centering
\includegraphics[scale=0.45]{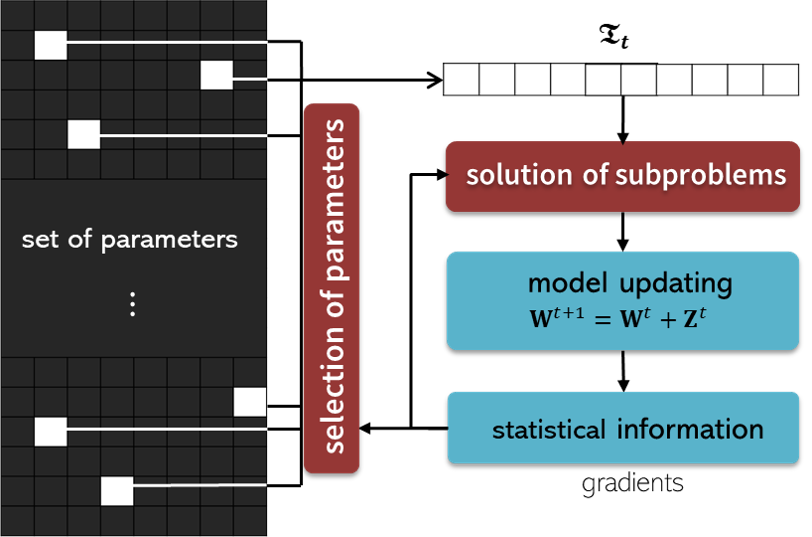}
\caption{\label{Fig:CGD}The diagram of coordinate gradient descent, where the steps in red are key points for improving computational efficiency.}
\end{figure}

\vspace{0.1cm}
\noindent
\textbf{Motivation.} We first present the basic formulation of CGD \cite{wright2015coordinate}. Let $Q{(\mathbf{W})}$ denote the objective function with the parameter matrix of $\mathbf{W}$, and ${\mathcal{I}_t}$ contain the indexes of the selected parameters in the $t$-th iteration. As only a few parameters need to be updated, CGD can take  all instances into account and optimize the selected parameters to optimal solutions in each iteration.  Specifically, let $\mathbf{W}+\mathbf{Z}^t$ refer to the expected solution on selected parameters, namely $\mathbf{Z}^t$ only contains non-zero values at the indexes in ${\mathcal{I}_t}$. In each iteration, CGD minimizes the following subproblem
\begin{eqnarray}\label{eq:CGD1}
g(\mathbf{Z}^t)= {Q}(\mathbf{W}+\mathbf{Z}^t)-{Q}(\mathbf{W}),
\end{eqnarray}
where $Z^t_{i,j}=0, \forall \{i,j\} \notin \mathcal{I}_t$.
Although the scales of the reduced subproblems are small and many available solvers can be directly used for the optimization \cite{akata2014good}, CGD can still lead to an expensive time cost for exact solutions during iterative optimization. Besides, updating each parameter with the same number of iterations causes a massive amount of redundant computations.

To address these issues, we review LML methods from two aspects: the selection of parameters and the fast solution of subproblems. For convenience, an illustrative diagram showing their roles is displayed in  Fig.\ref{Fig:CGD}.

\vspace{0.1cm}
\noindent
\textbf{Selection of parameters.} For convergence with a smaller number of iterations, the indexes must be selected carefully. Typically, the selection of these parameters follows a Gauss-Seidel manner \cite{young2014iterative}, namely, each parameter is updated at least once within a fixed number of consecutive iterations. As features may be correlated, performing the traversal with a random order of parameters in each iteration has empirically shown the ability of acceleration \cite{hsieh2008dual,nesterov2012efficiency,chang2008coordinate}. On the other hand, Gauss-Southwell (GS) rules propose to take advantage of gradient information for the selection, which can be regarded as performing the steepest descent \cite{shi2016primer}. To avoid the expensive cost in GS rules, we can connect them to the nearest neighbor search and introduce a tree structure to approximate the rules \cite{nutini2015coordinate,dhillon2011nearest}. Besides, by introducing the quadratic approximation on objective functions and performing the diagonal approximation on Hessian matrices, the selection of multiple indexes can be reduced to separable problems \cite{yun2011coordinate}.

\vspace{0.1cm}
\noindent
\textbf{Fast solutions of subproblems.} Similar to accelerated gradient descent, extrapolation steps can be employed for accelerated coordinate descent methods \cite{nesterov2012efficiency,lee2013efficient}. Suppose $Q(\mathbf{W})$=$p(\mathbf{W})$+$q(\mathbf{W})$, where $p(\mathbf{W})$ and $q(\mathbf{W})$ are smooth and nonsmooth, respectively. Accelerated proximal gradient (APG) first replaces $p(\mathbf{W})$ with the first-order approximation regularized by a trust region penalty and then uses the information at an extrapolated point to update the next iterate \cite{lin2014accelerated}. With appropriate positive weights, it can improve the convergence remarkably while remaining almost the same per-iteration complexity to the proximal gradient \cite{schmidt2011convergence,beck2009fast}.
To address nonconvex problems, the monotone APG method proposed in \cite{li2015accelerated} introduces sufficient descent conditions with a line search. It reduces the average number of proximal mappings in each iteration and speeds up the convergence. Moreover, if both  $p(\mathbf{W})$ and $q(\mathbf{W})$ are nonsmooth, ADMM introduces Douglas-Rachford splitting to solve the problem, which is extremely useful when the proximal operators can be evaluated efficiently  \cite{boyd2011distributed}. Of note, similar to standard iterative optimization, common heuristics can be used for speedup, such as caching variable-dependent eigendecomposition \cite{boyd2011distributed} and pre-computation of non-variable quantities \cite{shi2016primer}.

\vspace{0.1cm}
\noindent
\textbf{Discussion based on experimental results.} Now we discuss the efficiency of the above strategies for solving primal and dual problems. On the one hand, when optimizing L2-SVM with correlated features, the CGD based on random permutation could obtain the same relative difference to the optimum with 2$\times$ to 8$\times$ fewer training costs \cite{chang2008coordinate}. Besides, due to the sparsity of the solutions of L1 and L2 regularized least squares problems, CGD based on GS rules not only outperformed random and cyclic selections with 5$\times$ to 10$\times$ fewer epochs for faster convergence but also used 2$\times$ less running time including the estimation of statistical information \cite{nutini2015coordinate}. On the other hand, for solving lasso, APG only required 20\% time of the basic proximal gradient, and ADMM further saved 50\% time costs \cite{parikh2014proximal}. Besides, when solving linear inverse problems like image deblurring, the result of the proximal algorithm ISTA after 10,000 iterations could be obtained by its accelerated version with 275 iterations \cite{daubechies2004iterative,beck2009fast}. For nonconvex logistic regression with capped L1 penalties, nonmonotone APG with released conditions obtained a higher accuracy with 75\% fewer costs than monotone APG \cite{li2015accelerated}. In addition, by caching eigendecomposition, ADMM solved problems 20$\times$ faster than computing it repeatedly in each iteration \cite{boyd2011distributed}.

{\color{black}\subsubsection{For Numerical Integration with MCMC}
The methods based on Markov chain Monte Carlo (MCMC) are widely used in Bayesian posterior inference on $p(\mathbf{W}|\mathcal{X})$ \cite{blei2003latent}, such as M-H and Gibbs sampling \cite{chib1995understanding,griffiths2004finding}. Unlike the gradient-descent optimization with multi-dimensional integrals on $p(\mathcal{X})$=$\int_\mathbf{W}p(\mathcal{X}|\mathbf{W})p(\mathbf{W})d\mathbf{W}$, they introduce numerical approximations by recording samples from the chain and avoid the huge time costs in high-dimensional models. Specifically, they first generate candidate samples by picking from distributions and then accept or reject them based on the corresponding acceptance ratios. To enhance acceptance probabilities and improve efficiency, more methods introduce the gradient of the density of target distributions to generate samples at high-density regions.

\vspace{0.1cm}
\noindent
\textbf{Motivation}: Given a dataset of $n$ instances, most gradient-based sampling methods \cite{neal2011mcmc} introduce the Langevin dynamic and update parameters based on both the gradient and the Gaussian noise as
\begin{eqnarray}\label{eq:MCMC1}
\mathbf{W}_{t}+\frac{\epsilon_{t}}{2}{\{\partial_{\mathbf{W}_t}\mathrm{log}p(\mathbf{W}_t)+\sum_{x \in \mathcal{X}}\partial_{\mathbf{W}_t}\mathrm{log}p(x|\mathbf{W}_t)\}}+v_t,
\end{eqnarray}
where the step size $\epsilon_{t}$ and the variances of noise $v_t$$\sim$$ N(0,\epsilon_t)$ are balanced. As a result, their trajectories of parameters can converge to the full posterior distribution rather than just the maximum a posteriori mode. However, the calculation of gradients per interaction faces the computation over the whole dataset, and expensive M-H accept/reject tests are required to correct the discretization error. Although Gibbs sampling is free from M-H tests, it is still limited to the exact sampling from conditional posterior distributions. Thus, many stochastic gradient MCMC methods based on mini-batches are proposed to address the above issues.

\vspace{0.1cm}
\noindent
\textbf{Mini-batch gradient without M-H tests.} SGLD first introduces the scaled gradients that estimated from mini-batch instances to approximate the gradient of the log-likelihood \cite{welling2011bayesian}.  Besides, it discards the M-H test and accepts all the generated samples, since its discretization error disappears when $\epsilon_t$$\to$0. To reduce the number of iterations, SGFS prefers samples from approximated Gaussian distributions for large step sizes and switches to samples from the non-Gaussian approximation of posterior distributions for small step sizes \cite{ahn2012bayesian}. Meanwhile, SGHMC introduces a second-order Langevin dynamics with a friction term to improve the exploration of distant space \cite{chen2014stochastic}. In addition, SGRLD extends SGLD for models on probability simplices by using Riemannian geometry of parameter spaces \cite{patterson2013stochastic}, and SGRHMC further takes advantages of both the momentum of SGHMC and the geometry of SGLRD \cite{ma2015complete}.

\vspace{0.1cm}
\noindent
\textbf{Discussion based on experimental results.} Now we show the effectiveness of the above MCMC methods for Bayesian posterior inference. Firstly, these methods enable efficient inference for large-scale datasets. For example, SGRLD could perform LDA on Wikipedia corpus while Gibbs sampling was not able to run \cite{patterson2013stochastic}. Secondly, different mini-batch gradient variants can inherit the properties of their original versions. For example, SGHMC could generate high-quality distant samples and reduce more than 50\% iterations than SGLD for training Bayesian neural networks \cite{chen2014stochastic}. Besides, SGRHMC resulted in lower perplexities of LDA than both SGRLD and SGHMC while remaining similar costs \cite{ma2015complete}.}

\subsubsection{Summary}
To enhance the computational efficiency and remain reliable solutions, the above methods prefer the computations that produce higher reduction in optimization errors.  Firstly, both MGD and CGD methods take the importance of involved instances or parameters into account and introduce the careful selections for reducing the time cost. Meanwhile, both can take advantage of Nesterov's extrapolation steps and tune learning rates with accumulated gradients or use a line search for faster convergence. Secondly, CGD is more suitable for optimizing models with a large number of parameters, especially for linear models where data is stored with inverted indexes. In contrast, if the number of features is much smaller than the number of instances, one should solve problems based on MGD. Thirdly, for handling large models with a massive amount of instances, it is natural to combine two strategies together, i.e., updating a subset of parameters with a few instances in each iteration. For example, a recently proposed method called mini-batch randomized block coordinate descent estimates the partial gradient of selected parameters based on a random mini-batch in each iteration \cite{zhao2014accelerated}. {\color{black}Finally, mini-batch gradient MCMC algorithms improve qualities of samples and accelerate Bayesian inference. Besides, these algorithms generally resemble MGD from different views, such as additive noise and momentum. However, since samples from the dynamics of native stochastic variants may not always converge to the desired distribution, some necessary modification terms must be introduced to substitute the correction steps \cite{chen2014stochastic,ma2015complete}. Besides, the theory on how these additions differentiate a Bayesian algorithm from its optimization remains to be studied \cite{chen2016bridging}.}

Meanwhile, a significant part of the above LML methods focuses on handling convex problems, while only a few methods can address nonconvex problems. Besides, most methods solve target problems directly. Instead, one may seek the transformed forms of their problems and develop more appropriate algorithms by referring related work in other domains. Details will be discussed in Sec.4.1.2.

\begin{table*}[tbp]
\centering
\caption{\label{Tab:parallel}A brief lookup table for LML methods based on Computation Parallelism.}
\linespread{5}
\begin{tabular}{lll}
\hline\hline
 \textbf{Categories}       &  \multirow{1}{*}{\textbf{Strategies}}& \multirow{1}{*}{\textbf{Representative Methods}} \\ \hline
\multirow{1}{*}{Multi-Core}  &  adoption of optimized libraries & acceleration with CPUs \cite{chiang2016parallel,chin2015fast,sonnenburg2010shogun,igel2008shark}, with GPUs \cite{bergstra2011theano,jia2014caffe,abadi2016tensorflow,paszke2019pytorch}.  \\ \cline{2-3}
Machines & improvement  of  i/o  accesses & balance memory and disks \cite{kyrola2012graphchi,roy2013x,zhu2015gridgraph}, balance CPU and GPU memory \cite{rhu2016vdnn}. \\ \cline{1-3}
\multirow{1}{*}{ } & mapreduce abstraction & mapreduce \cite{shvachko2010hadoop,rosen2013iterative}, \cite{boehm2014hybrid}, spark \cite{Zaharia2010Spark}, supporting libraries \cite{meng2016mllib:,zaharia2013discretized,boehm2016systemml}. \\ \cline{2-3}
Multi-Machine &graph-parallel abstraction& bulk synchronous  \cite{malewicz2010pregel}, asynchronous  \cite{low2012distributed}, GAS decomposition \cite{gonzalez2012powergraph,chen2019powerlyra}. \\ \cline{2-3}
Clusters&parameter server abstraction& distributed networks \cite{dean2012large,Chen2015MXNet}, gradient boosting \cite{jiang2018dimboost}, general-purpose system \cite{xing2015petuum}.  \\ \cline{2-3}
&allreduce abstraction& tree-based \cite{chen2016xgboost}, butterfly-based \cite{ke2017lightgbm,distributedlightgbm}, ring-based \cite{sergeev2018horovod,jeaugey2017nccl,distributedPYTORCH,distributedtensorflow}, hybrid \cite{goyal2017accurate}.	\\ \hline	\hline		
\end{tabular}
\end{table*}

\subsection{Computation Parallelism}
Computation parallelism reduces practical time costs from the perspective of computational capabilities. The motivation is, mutually-independent subtasks can be processed simultaneously over multiple computing devices. For this purpose, many parallel processing systems have been developed. On the one hand, these systems increase the number of computations per time unit based on powerful hardware; on the other hand, they introduce advanced software to schedule hardware efficiently. According to the scenarios of hardware, we review the systems from two categories, as shown in Tab.\ref{Tab:parallel}, including the systems with multi-core machines and those with multi-machine clusters. Below we introduce their main characteristics, aiming at utilizing their developments to benefit efficient machine learning methods.

\subsubsection{For Multi-core Machines}
The systems here take advantage of multi-core machines for parallelism. Specifically, all cores can access common memories, and each of them execute subtasks independently.

\vspace{0.1cm}
\noindent
\textbf{Motivation.}
To enhance the computational ability of a single machine, recent processing systems increase the number of parallelable instructions by introducing multiple processors into the machine, such as multiple-core CPUs and GPUs. Although their theoretical capabilities could be improved almost linearly with respect to the number of cores, the practical speedup is much lower owing to the overhead. To improve the effectiveness, users have to handle tedious works such as the scheduling of cores and memory.

To address this issue, we review representative works that provide the interfaces with different levels of simplicity and flexibility. Consequently, one can take their advantages and smoothly implement LML models in a single machine.

\begin{figure*}[tbp]
\centering
\includegraphics[scale=0.6]{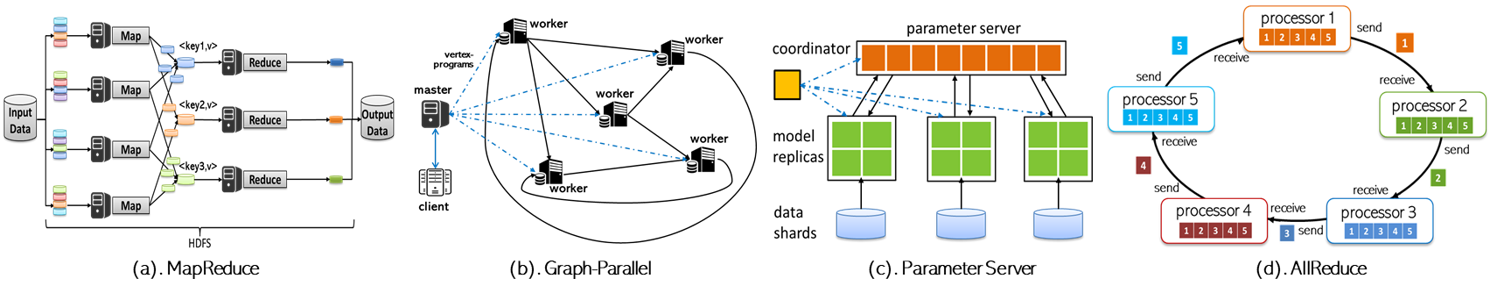}
\caption{\label{Fig:DisCom}The toy examples of distributed processing systems. (a) the structure of a mapreduce system \cite{dean2008mapreduce}, (b) the structure of graph-parallel systems \cite{malewicz2010pregel}, (c) a sketch of parameter-server-based cluster \cite{li2014scaling}, and (d) a step of data communication of the ring-based allreduce abstraction \cite{sergeev2018horovod}.}
\end{figure*}

\vspace{0.1cm}
\noindent
\textbf{Adoption of highly-optimized libraries.}
When the memory is sufficient, we can directly utilize multi-core CPUs or GPUs to perform the optimization with available highly-optimized libraries, e.g., Mkl for Intel CPUs and Cuda for Nvidia GPUs. On the one hand, CPUs can use relatively cheap RAMs and handle complex operations in parallel. For example, Liblinear and FPSG provide the efficient solutions to L1-regularized linear classification and parallel matrix factorization \cite{chiang2016parallel,chin2015fast}, respectively. For general purposes, {Shogun} is compatible with support vector machines and multiple kernel learning \cite{sonnenburg2010shogun}. {Shark} additionally supports evolutionary algorithms for multi-objective optimization \cite{igel2008shark}. Of note, lock-less asynchronous parallel (ASP) can further take advantage of CPU cores and reduce the overhead for parallelizing MGD \cite{recht2011hogwild}. On the other hand, GPUs built upon thousands of stream cores are better at handling simple computations in parallel. In particular, deep learning systems are highly benefited from GPU-accelerated tensor computations and generally provide automatic differentiation, such as Theano \cite{bergstra2011theano}, {Caffe} \cite{jia2014caffe}, {MXNet} \cite{Chen2015MXNet}, {TensorFlow} \cite{abadi2016tensorflow} and {PyTorch} \cite{paszke2019pytorch}. For example, {TensorFlow} optimizes execution phase based on the global information of programs and achieves the high utilization of GPUs \cite{abadi2016tensorflow}. {PyTorch} introduces easy-to-use dynamic dataflow graphs and offers the interface to wrap any module to be parallelized over batch dimension \cite{paszke2019pytorch}.

\vspace{0.1cm}
\noindent
\textbf{Improvement of I/O accesses.} When the memory cannot handle a LML model at once, we have to break it into multiple parts, where each part is computed in parallel and different parts are implemented sequentially. In this case, the I/O accesses result in an in-negligible overhead for overall computations, and the effective utilization of memory and extra storage becomes essential. For example, to handle a large graph with $n_e$ edges, GraphChi divides its vertices into many intervals, where each is associated with a shard of ordered edges \cite{kyrola2012graphchi}. When performing updating from an interval to the next, it slides a window over each of the shards for reading and writing, implementing the asynchronous model with $O(2n_e)$ accesses. To avoid expensive pre-sorting of edges, X-Stream instead proposes an edge-centric implementation in synchronous by streaming unordered edge lists \cite{roy2013x}. However, it needs to read edges and generate updates in the scatter phase and read updates in the gather phase, leading to $O(3n_e)$ accesses. Recently, GridGraph introduces a two-level partitioning, which streams every edge and applies the generated update instantly \cite{zhu2015gridgraph}. It only requires one read pass over edges and several read/write passes over vertices, leading to nearly $O(n_e)$ accesses. Considering that the memory of GPU is expensive and intermediate feature maps account for the majority of usage, vDNN moves the intermediate data between GPU and CPU, which can train larger networks beyond the limits of the GPU's memory \cite{rhu2016vdnn}.

\vspace{0.1cm}
\noindent
\textbf{Discussion based on experiments.}
The systems based on multi-core machines have shown their power to accelerate the implementation of LML. In particular, when the memory is sufficient, FPSG could speed up matrix factorization nearly 7$\times$ with a 12-core CPU \cite{chin2015fast}. Besides, Theano was able to achieve 5$\times$ faster for training CNNs on GPUs rather than CPUs and 6$\times$ faster for optimizing DBNs \cite{bergstra2011theano}. Even with limited memory, by scheduling I/O accesses efficiently, we can still utilize the parallelism and train large models within acceptable time costs. For example, GraphChi used a mac mini and successfully solved large Pagerank problems reported by distributed systems \cite{kyrola2012graphchi}. For the same number of iterations, GridGraph only required fewer than 50\% time costs of others \cite{zhu2015gridgraph}. Besides, with the support of CPU memory, vDNN could optimize a deep network with the memory requirement of 67 GB based on a 12GB GPU \cite{rhu2016vdnn}.

\subsubsection{For Multi-machine Clusters}
With the explosion of data size, processing systems based on distributed clusters have attracted increasing attention. These systems utilize local area networks to integrate a set of machines, where each runs its own tasks.

\vspace{0.1cm}
\noindent
\textbf{Motivation.} Distributed processing systems introduce parallelism at two levels, including a number of nodes in a cluster and multiple cores within the individual node. One can also construct huge models for complex tasks if model parallelism is applied. Nevertheless, the main challenge of applying a cluster is managing (heterogeneous) machines, such as data communication across the nodes. These tedious works may lead to low resource utilization.

Below we review typical distributed systems to simplify the developments of parallelism \cite{cai2014comparison}. We present the introduction according to their top-level topology configurations.

\vspace{0.1cm}
\noindent
\textbf{MapReduce abstraction.} The MapReduce abstraction divides computations among multiple machines \cite{dean2008mapreduce}, where each works with a part of tasks in parallel with its locality-distributed data. As we can see from Fig.\ref{Fig:DisCom}(a), MapReduce systems reformulate complex models into a series of simple Map and Reduce subtasks. Specifically,  {Map} transforms its local data into a set of intermediate key/value pairs, and {Reduce} merges all intermediate values associated with the same intermediate key. The systems take care of the details of data partitioning, execution scheduling, and communication managing. As a result,  the methods that meet this abstraction can take advantage of locality-distributed data and  be executed on clusters smoothly.

{Hadoop MapReduce} is the first open-source implementation, which introduces HDFS files to store the data over disks \cite{shvachko2010hadoop}. Its flexibility allows new data-analysis software to either complement or replace its original elements, such as YARN \cite{vavilapalli2013apache}. Besides, to support iterative machine learning methods, {Iterative MapReduce} introduces a loop operator as a fundamental extension \cite{rosen2013iterative}, and {Hybrid MapReduce} presents a cost-based optimization framework to combine both the task and data parallelism \cite{boehm2014hybrid}. On the other hand, {Spark} employs the immutable distributed collection of objects RDD as its architectural foundation \cite{Zaharia2012Resilient} and distributes the data over a cluster in the memory to speed up iterative computations \cite{Zaharia2010Spark}. Besides, a stack of libraries are developed to further simplify its programming. For example, {MLlib} provides a variety of linear algebra and optimization primitives \cite{meng2016mllib:} and {Spark Streaming} eases the streaming learning \cite{zaharia2013discretized}. Recently, SystemML enables Spark to compile sophisticated machine learning methods into efficient execution plans automatically \cite{boehm2016systemml}.

\vspace{0.1cm}
\noindent
\textbf{Graph-parallel abstraction.} The graph-parallel abstraction shown in Fig.\ref{Fig:DisCom}(b) exploits the structure of sparse graphs to improve communication between different machines \cite{kalavri2018high}. Specifically, it consists of a sparse graph and a vertex-program. The former is partitioned into different subsets of vertexes over multiple machines. The latter is executed in parallel on each vertex and can interact with the neighboring vertexes with the same edge. Compared with MapReduce, these systems constrain the interaction based on sparse adjacency and reduces the cost of communication.

{Pregel} is a fundamental graph-parallel system running in bulk synchronous parallel (BSP) \cite{malewicz2010pregel}. It consists of a sequence of supersteps, where messages sent during one superstep are guaranteed to be delivered at the beginning of the next superstep. However, since the slowest machine determines the runtime in each iteration, {Pregel} can result in idling problems. {GraphLab} thus introduces ASP by releasing the constraints of supersteps \cite{low2012distributed}, which updates the vertexes using the most recent values. Moreover, {PowerGraph} introduces a gather, apply, and scatter (GAS) decomposition to factor vertex-programs for power-law graphs \cite{gonzalez2012powergraph} and enables the computations in both BSP and ASP. Recently, PowerLyra uses centralized computation for low-degree vertices to avoid frequent communications and follows GAS to distribute the computation for high-degree vertices \cite{chen2019powerlyra}. Similarly, it supports both BSP and ASP.

\vspace{0.1cm}
\noindent
\textbf{Parameter server abstraction.} The parameter server abstraction distributes data and workloads over worker nodes and maintains distributed shared memory for parameters on server nodes \cite{li2014scaling}. An example is shown in Fig.\ref{Fig:DisCom}(c). This abstraction  enables an easy-to-use shared interface for I/O access to models, namely, workers can update and retrieve the different parts of parameters as needed.

DistBelief first utilizes computing clusters with the parameter server \cite{dean2012large}, where the model replicas in workers asynchronously fetch parameters and push gradients with the parameter server. It thus can train deep networks with billions of parameters using thousands of CPU cores. After that, {MXNet} adopts a two-level structure to reduce the bandwidth requirement \cite{Chen2015MXNet}, where one level of servers manages the data synchronization within a single machine, and the other level of servers manages intermachine synchronization. Moreover, the naive distributed Tensorflow introduces dataflow with the mutable state to mimic the functionality of a parameter server, which provides additional flexibility on optimization algorithms and consistency schemes \cite{abadi2016tensorflow}. Besides, to handle GBDT with high dimensions, DimBoost transforms each local histogram to a low-precision histogram before sending it to the parameter server \cite{jiang2018dimboost}. Once these local histograms are merged on the servers, a task scheduler assigns active tree nodes among all workers to calculate the best split for each tree node. {Meanwhile, distributed MCMC methods are also developed for large-scale Bayesian matrix factorization \cite{ahn2015large}, where the parameter server updates its global copy with new sub-parameter states from workers.} In addition, as a general-purpose system, Petuum is also benefited from this abstraction \cite{xing2015petuum} and simplifies the distributed development of various scalable machine learning models.

\vspace{0.1cm}
\noindent
\textbf{Allreduce abstraction.} {The allreduce abstraction reduces the target arrays in all $m$ machines to a single array and returns the resultant array to all machines. It retains the convergence of gradient descent in BSP and can be divided into tree-based \cite{chen2015rabit}, butterfly-based \cite{wu2015deep}, and ring-based \cite{baiduallreduce}.

XGBoost introduces tree-allreduce by organizing all the workers as a binomial tree \cite{chen2016xgboost}. Its aggregation steps follow a bottom-to-up scheme starting from the leaves and ending at the root. However, these steps cannot overlap in its implementation. LightGBM thus introduces butterfly-allreduce with a recursive halving strategy \cite{ke2017lightgbm,distributedlightgbm}. Specifically, at the $k$-th step, each worker exchanges $\frac{n}{{2k}}$ data with a worker that is $\frac{m}{{2k}}$ distance away. This process iterates until the nearest workers exchange their data.

Recently, ring-allreduce has attracted more attention by employing bandwidth-optimal algorithms to reduce communication overhead \cite{patarasuk2009bandwidth}. Fig.\ref{Fig:DisCom}(d) displays an illustrative example, where $m$ machines communicate with their neighbors 2$\times$($m$-1) times by sending and receiving parameter buffers. Specifically, in the first (second) $m$-1 iterations, each machine receives a buffer and adds it to (substitutes it for) its own value at the corresponding location. These new values will be sent in the next iteration. To ease the use of ring-allreduce on GPUs, Nvidia develops Nccl for its devices \cite{jeaugey2017nccl}. After that, Horovod integrates Nccl into deep learning and allows users to reduce the modification of their single-GPU programs for the distributed implementation \cite{sergeev2018horovod}. Pytorch utilizes Nccl to operate heavily-parallel programs on independent GPUs  \cite{distributedPYTORCH}. Recently, Tensorflow also provides two different experimental implementations for ring-allreduce \cite{distributedtensorflow}. 
In particular, Caffe2 introduces butterfly-allreduce for inter-machine communication while using ring-allreduce for local GPUs within each machine \cite{goyal2017accurate}.

\vspace{0.1cm}
\noindent
\textbf{Discussion based on experiments.} Effective distributed systems not only allow us to handle larger-scale datasets and models, but also make data communication more efficient. However, different abstractions show their characteristics in specific tasks. For example, for PageRank, Hadoop and Spark took 198 and 97 seconds for each iteration, while PowerGraph only used 3.6 seconds and was significantly faster \cite{gonzalez2012powergraph}. Based on the differentiated partition for low-degree and high-degree vertices, PowerLyra further outperformed PowerGraph with a speedup ranging from 1.4$\times$ to 2$\times$. For matrix factorization, Petuum was faster than Spark and GraphLab \cite{xing2015petuum}. Besides, when the number of workers was not a power of two, DimBoost could outperform XGBoost and LightGBM significantly by merging parameters with data sparsity \cite{jiang2018dimboost}. Based on ring-allreduce, Horovod makes TensorFlow more scalable on a large number of GPUs. In particular, it could half the time cost of naive distributed Tensorflow when training CNNs with 128 GPUs \cite{sergeev2018horovod}.

\subsubsection{Summary}
Now we provide rough guidance of parallelism strategies.
Firstly, CPUs are optimized for handling the models with complex sequential operations. For example, by introducing locality sensitive hashing for the sparse selection of activate neurons, the training of FCNs on CPUs was 3.5$\times$ faster than the best available GPUs \cite{chen2019slide}. On the contrary, GPUs could enhance the computational capability greatly for the intensive computational models with huge numbers of dense matrix and vector operations \cite{fatahalian2004understanding}. Secondly, although machine learning methods may be compatible with multiple abstractions \cite{ooi2015singa}, it is crucial to choose a proper abstraction for a specific setting. In general, graph-parallel and ring-allreduce are more suitable for graph models and deep learning models, respectively. Thirdly, we can ease the development of higher-level systems by leveraging the existing systems. For example, built upon YARN, Angel develops a parameter server system for general-purpose LML \cite{jiang2018angel}, which supports both model and data parallelism. Similarly, Glint applies Spark for convenient data processing in memory and introduces an asynchronous parameter server for large topic models \cite{jagerman2017computing}. Moreover, PS2 launches Spark and parameter servers as two separate applications without hacking Spark \cite{zhang2019ps2}, which makes it compatible with any version of Spark. {GraphX} also introduces the graph-parallel to accelerate graph-based models on Spark \cite{gonzalez2014graphx}.

\begin{table*}[tbp]
\centering
\caption{\label{Tab:hybrid}A brief lookup table for LML methods based on Hybrid Collaboration.}
\linespread{5}
\begin{tabular}{lll}
\hline\hline
 \textbf{Categories}&  \multirow{1}{*}{\textbf{Strategies}}& \multirow{1}{*}{\textbf{Representative Methods}} \\ \hline
\multirow{2}{*}{Gradient}  &  Compression &  quantize component values \cite{seide20141,alistarh2017qsgd,wen2017terngrad,jiang2018sketchml,zhang2017zipml}, remove redundancy gradients \cite{lin2017deep}. \\ \cline{2-3}
 & Delay & limited delay with stale synchronous  \cite{langford2009slow,agarwal2011distributed,jaggi2014communication,richtarik2016distributed,zhang2020deep,zheng2017asynchronous}, infinite delay \cite{zinkevich2010parallelized}.
\\ \hline	\hline		
\end{tabular}
\end{table*}

However, the best choice of practical developments varies for the size of datasets and need further studies. For example, for training SVMs, the efficiency of GPUs only became higher than CPUs when the sizes of datasets increased to some degree \cite{li2011fast}. Besides, owing to the overhead, the return of distributed systems becomes lower with the increasing number of processors, and we have to control the return on investment regardless of abstractions. For example, although larger CNNs was supposed to benefit more with multiple machines, a model with 1.7 billion parameters only had 12$\times$ speedup using 81 machines \cite{dean2012large}.
}

\subsection{Hybrid Collaboration}
The methods in the above sections scale up LML from the perspective of computational complexities, computational efficiency, and computational capabilities, respectively. Thus, they are independent of each other, and any partial enhancement enables the overall improvement.

{
\vspace{0.1cm}
\noindent
\textbf{Motivation.} In general, we can enhance machine learning efficiency by jointly using multiple strategies directly \cite{calandriello2018improved,dunner2018snap,grave2017efficient}. The most typical work is deep learning, which has been undergoing unprecedented development over the past decades \cite{lecun2015deep,chen2014bigA,chen2014bigB}. The key reasons that facilitate its rapid development are detailed as follows. Firstly, deep learning methods introduce prior knowledge such as filter decomposition to simplify predictive models with fewer parameters \cite{ying2018hierarchical}. Secondly, owing to its mini-batch-based optimization, various advanced MGD algorithms can be employed to train neural networks, leading to faster converges \cite{Kingma2014Adam}. Thirdly, since GPUs are paired with a highly-optimized implementation of 2D convolutions and well-suited to cross-GPU parallelization\cite{krizhevsky2012imagenet}, it is efficient for researchers to debug their advanced networks \cite{yu2018gradiveq}.

However, the above perspectives can be further improved for synergy effects. In particular, for simplified models, we can modify optimization algorithms to improve parallelism. We divided them into two strategies as listed in Tab.\ref{Tab:hybrid}, including gradient compression and gradient delay.

\vspace{0.1cm}
\noindent
\textbf{Gradient compression.} {\color{black}To further accelerate the data communication in distributed optimization, many compression methods have been proposed}, where only a small number of data is communicated across machines. For example, 1-bit MGD quantizes each component of gradients aggressively to one bit per value and feeds back the quantization error across mini-batches \cite{seide20141}. On the contrary, QSGD directly quantizes the components by randomized rounding to a discrete set of values while preserving the expectation with minimal variance \cite{alistarh2017qsgd}. As a result, it makes a trade-off between the number of bits communicated per iteration and the added variance. To exploit valuable gradients with outliers, TernGrad introduces layer-wise ternarizing and gradient clipping \cite{wen2017terngrad}. Meanwhile, DGC only transmits gradients larger than a threshold for sparsification while accumulating the rest of gradients locally \cite{lin2017deep}. In particular, when gradients are sparse and their distribution is nonuniform, the nonzero elements in a gradient are generally stored as key-value pairs to save space. SketchML thus takes the distribution into account and uses a quantile sketch to summarize gradient values into several buckets for encoding \cite{jiang2018sketchml}. Besides, it stores the keys with a delta format and transfers them with fewer bytes. Furthermore, ZipML proposes to optimize the compression on gradients, models, and instances and then perform an end-to-end low-precision learning \cite{zhang2017zipml}. Of note, the above compression methods are compatible with various parallel systems, such as multiple GPUs \cite{seide20141,alistarh2017qsgd} and parameter servers \cite{wen2017terngrad,jiang2018sketchml,lin2017deep}.

\vspace{0.1cm}
\noindent
\textbf{Gradient delay.} ASP-based MGD and CGD algorithms  \cite{recht2011hogwild,liu2014asynchronous} minimize the overhead of locking and alleviate the idling issues. However, they remain huge costs of data communication. The stale synchronous parallel (SSP) algorithms with gradient delays thus get more attention \cite{ho2013more}. For example, \cite{langford2009slow} orders CPU cores where each updates variables in a round-robin fashion. Its delay reduces the cost of reading parameters of the latest models. \cite{agarwal2011distributed} introduces tree-allreduce in distributed settings, where each parent averages the gradients of the children nodes from the previous round with its own gradient, and then passes the result back up the tree. In addition, CoCoA and Hydra use workers to perform some steps of optimization with their local instances and variables in each iteration \cite{jaggi2014communication,richtarik2016distributed}, respectively. EASGD introduces an elastic force between central and local models for more exploration to avoid many local optima \cite{zhang2020deep}. Moreover, \cite{zinkevich2010parallelized} proposes to solve subproblems exactly on each machine without communication between machines before the end, namely infinite delay. Since the delayed gradient is just a zero-order approximation of the correct version, DCASGD leverages the Taylor expansion of gradient functions and the approximation of the Hessian matrix of the loss function to compensate for  delay \cite{zheng2017asynchronous}. On the other hand, delayed gradients can also be utilized to improve the convergence, which in turn reduces the communication. For example, ECQSGD and DoubleSqueeze accumulate all the previous quantization errors rather than only the last one for error feedback \cite{wu2018error, tang2019doublesqueeze}. Besides, \cite{mcmahan2014delay} introduces the delayed gradients to chooses adaptive learning rates in parallel settings.

\vspace{0.1cm}
\noindent
\textbf{Discussion based on experiments.} Now we give a brief discussion on the methods of hybrid collaboration. Firstly, gradient compression can reduce not only the time cost per iteration but also the overall cost. For example, compared with full 32-bit gradients, the training of 16-GPU AlexNet with 4-bit QSGD led to more than 4$\times$ speedup on communications, and 2.5$\times$ speedup on overall costs for the same accuracy \cite{alistarh2017qsgd}. In particular, for linear models, SketchML could accelerate the original optimizer and achieve more than 4$\times$ improvements \cite{jiang2018sketchml}. Of note, models with larger communication-to-computation ratios and networks with smaller bandwidth can benefit more from compression \cite{wen2017terngrad}.
Secondly, gradient delay also reduces the amount of data communication and remarkably increases the proportion of time workers spend on the computation. For example, based on local updates, CoCoA was able to converge to a more accurate solution with 25$\times$ faster than the best non-locally updating competitor \cite{jaggi2014communication}. Besides, the reduction of communication cost could be nearly linear with respect to the delay under a large number of workers \cite{zhang2020deep}.
}

{\section{Discussions}}
Existing LML methods have established a solid foundation for big data analysis. Below we outline some promising extension directions and discuss important open issues.

\vspace{0.5cm}
\subsection{Extension Directions}
\subsubsection{For Model Simplification}
The methods of model simplification reduce computational complexities with adequate prior knowledge. Therefore, we may further explore the distributions and structures of instances to enhance the scalability of predictive models.

{
\vspace{0.1cm}
\noindent
\textbf{Explore distributions of data.} The low-rank approximation for kernels assumes that mapped instances in RKHS nearly lie on a manifold of a much lower dimension, and their inner products can be estimated on the low-dimension space. Graph-based methods are also scaled up based on the assumption that nearby points are more likely to share labels. Thus, to make use of large-scale unstructured data, the underlying  global and local data distributions must be considered \cite{gilardi2000local}. Based on this motivation, HSE and AER build hierarchical indexes on unlabeled instances for coarse-to-fine query selection \cite{Aodha2014Hierarchical,fu2018scalable}. As a result, although the cost of the quality estimation of each candidate is not reduced, they lower computational complexities significantly by decreasing the number of candidates.

\vspace{0.1cm}
\noindent
\textbf{Exploit structures within instances.} Various types of convolutions based on filter decomposition have been proposed to reduce the sizes of deep models. These works demonstrate that to make receptive fields more effectively, a critical direction is to take the features with multi-scale invariance into account and make use of the structural information of instances. Following this motivation, DCNs thus develop deformable convolutions to model larger transformations efficiently \cite{dai2017deformable,zhu2019deformable}, which only add a few parameters.}

\vspace{0.1cm}
\noindent
\textbf{Analyze objective tasks.}
A more in-depth analysis of objective tasks provides another powerful way for model simplification. For example, inspired by the coarse-to-fine categorization of visual scenes in scene-selective cortex \cite{musel2014coarse}, hierarchical convolutional networks are developed for image classification \cite{yan2015hd}. These networks decrease the number of parameters and correspondingly reduce the involved computations per iteration. In addition, distributing parameters and computations according to the frequencies of objective predictions also reduces computational complexities \cite{mikolov2013distributed}.

\subsubsection{For Optimization Approximation}
To enhance computational efficiency, the methods of optimization approximation prefer the computations that bring significant reduction in optimization errors. However, most of them focus on convex problems. Besides, researchers generally follow stereotyped routines to address their emerging problems without seeking more appropriate solutions. To further speed up the optimization for a broader range of scenarios, the following extensions can be considered.

\vspace{0.1cm}
\noindent
\textbf{Optimization for nonconvex problems.}
The optimization may fall into poor local minima when problems are nonconvex \cite{sun2019survey}, leading to unstable performances in real-world tasks. Therefore, it is necessary to enhance the ability of algorithms to escape or bypass the poor minima. To this end, various randomized operations can be considered, such as randomized weight initialization and random noise on gradients \cite{sutskever2013importance}. Besides, inspired by curriculum learning \cite{khan2011humans}, we may first utilize instances that are easy to fit and then gradually turn to difficult ones, such as boundary instances. As a result, it allows algorithms to obtain a good solution at early stages and then tune it for a better minimum.

\vspace{0.1cm}
\noindent
\textbf{Solutions inspired by mathematical models.}
By transforming target problems into classical models in other mathematical areas, such as geometry, we can borrow the experiences of handling these models to accelerate the optimization. For example, a simple iterative scheme used to find the minimum enclosing ball (MEB) can be used to solve large-scale SVM problems \cite{tsang2007simpler}. The motivation is, the dual of both MEB and SVM problems can be transformed into the same form. Similarly, \cite{joachims2006training} introduces cutting-plane algorithms that iteratively refine a feasible set to solve SVM on high-dimension sparse features.

\subsubsection{For Computation Parallelism}
Computation parallelism enhances computational capacities based on multiple computing devices. Benefit from existing systems, we can take advantage of flexible programming interfaces and low data communication for parallel data processing. However, with the fast upgrading of hardware, software, and predictive models, the following issues recently attract much attention.

\vspace{0.1cm}
\noindent
\textbf{Flexible hardware abstractions.} Highly-efficient parallel computations require fast data communication between different nodes. Although allreduce is able to minimize the communication overhead for most methods, it supposes that the memory and the computational ability of different nodes are at the same level. Otherwise, the utilization of resources will be limited by the weakest one. Considering the real-world scenarios where distributed systems generally consist of heterogeneous devices, more flexible abstractions need to be studied. A possible solution is to utilize domain-specific knowledge such as intra-job predictability for cluster scheduling \cite{peng2016towards}, \cite{xiao2018gandiva}. Besides, we may further explore hybrid abstractions and take account of traffic optimization to balance computational resources \cite{chen2018auto,goyal2017accurate}.

\vspace{0.1cm}
\noindent
\textbf{Modularized open-source software.} There are increasing interests in supporting open-source projects. Several motivations are as follows. The developments of parallel systems on general-purpose machine learning can be challenging even for large companies. Besides, open-source software can benefit more communities and democratize data science. To this end, normalized and modularized learning frameworks (similar to computer architectures \cite{von2012computer}) are preferred. As a consequence, any localized enhancement can be achieved by contributors, improving the overall efficiency. In addition, to construct a vibrant and lively community, it is necessary to provide easy-to-use interfaces with different programming languages to benefit more researchers \cite{moritz2018ray}, as well as the simple postprocessing for productization.

\subsection{Open Issues}
\subsubsection{Tighter Bounds of Complexities}
Various complexities during the procedure of data analysis can be used to evaluate required computational resources in real-world tasks. Below we discuss some important ones.

\vspace{0.1cm}
\noindent
\textbf{Bounds of computational complexities.}
The computational complexity determines the numbers of computing devices and the spaces of memory. However, many existing bounds of computational complexities only provide general guarantees for any probability distribution on instances and for any optimization algorithm that minimizes disagreement on training data. To estimate the complexities more precisely, it is necessary to establish tighter bounds for real-world data distributions and specific optimization algorithms, which may be achieved with the assistance of heuristical arguments from statistical physics \cite{agarwal2015Lower}.

\vspace{0.1cm}
\noindent
\textbf{Other related complexities.}
The measures of many other complexities also need to be studied. For example,  over-parameterized networks can be well trained with a much smaller set of instances than the one estimated based on the current sample complexity \cite{soltanolkotabi2017theoretical}. Besides, more instances are required for adversarial training to build robustness  \cite{schmidt2018adversarially}. In addition, the complexity of communication and the number of gates in a circuit can be used to estimate the practical limits on what machines can and cannot do \cite{arjevani2015communication}.

\subsubsection{Collection of valuable data}
Large-scale datasets play a crucial role in LML, as the patterns for making decisions are learned from data without being explicitly programmed. Therefore, our machine learning community is in great need of large-scale instance-level datasets, and efficient methods for generating such data either in a supervised or unsupervised manner. To this end, we highlight the following directions.

\vspace{0.1cm}
\noindent
\textbf{Annotation tools.} Annotation tools aim at building large-scale annotated datasets by collecting contributions from a lot of people, such as LabelMe in computer vision \cite{torralba2010labelme}. To reduce the costs of obtaining high-quality labels and simplify the annotation procedure, it is essential to ease the modes of annotation and interaction for oracles with scalable active learning methods \cite{joshi2012scalable}. Besides, it is difficult to collect clean instances with categorical and strong supervision information. The reasons are as follows. Firstly, there exist a large number of instances whose classes are fuzzy themselves. The one-hot or multi-hot encoding that groups them into specific classes can significantly harm their supervision information. Secondly, the inconsistent and inaccurate labels are ubiquitous in real-world situations due to subjective data-labeling processes. To address these issues, efficient weakly-supervised learning recently obtains urgent attention \cite{zhou2017brief}.

\vspace{0.1cm}
\noindent
\textbf{Data augmentation.} Data augmentation applies transformations to training sets to increase their scales. Based on the augmented data, we can improve the performance of predictive models, especially when original datasets are imbalanced or their instances are insufficient. These transformations can be as simple as flipping or rotating an image, or as complex as applying generative adversarial learning. Recently, BigGAN augments image datasets successfully by synthesizing high fidelity natural instances \cite{brock2018large}.

\subsubsection{Moderate Specialization of Hardware}
Over the past years, the design of computer architectures focuses on computation over communication. However, with the progress of semiconductor, we observed a new reality that data communication across processors becomes more expensive than the computation. Besides, the flexibility of general-purpose computing devices also makes the computation energy-inefficient in many emerging tasks. Although some special-purpose accelerators have been employed to be orders of magnitude improvements such as GPUs, FPGA, and TPUs, most of them are customized to a single- or narrow-type of machine learning methods. Therefore, for future hardware, it is necessary to exploit both the performance and energy-efficiency of specialization while broadening compatible methods \cite{hill201621st}.

\subsubsection{Quantum Machine Learning}
Quantum computers exploit superposition and entanglement principles of quantum mechanics, obtaining an immense number of calculations in parallel. Compared to digital computers, they can reduce both execution time and energy consumption dramatically, such as IBM-Q. However, since the programming models of these quantum computers are fundamentally different from those of digital computers, more attention should be paid to redesign the corresponding machine learning methods \cite{biamonte2017quantum,rebentrost2014quantum}.

\subsubsection{Privacy Protection}
With the development of LML, increasing importance has been attached to privacy protection. Specifically, when personal and sensitive data are analyzed in the cloud or other distributed environments \cite{agarwal2018cpsgd}, it is necessary to ensure that the analysis will not violate the privacy of individuals. Recently, federate learning provides an alternative by bringing codes to edge devices and updating the global model with secure aggregation \cite{bonawitz2019towards,bonawitz2017practical}.

\section{Conclusions}
Large-scale machine learning (LML) has significantly facilitated the data analysis in a mass scale over the past decades. However, despite these advances, the current LML requires further improvements to handle rapidly increasing data. In this paper, we first surveyed existing LML methods from three independent perspectives, namely, model simplification to reduce the computational complexity,  optimization approximation to improve computational efficiency, and computation parallelism to enhance the computational capability. After that, we discussed the limitations of these methods and the possible extensions that can make further improvements. Besides, some important open issues in related areas are presented. We hope that this survey can provide a clean sketch on LML, and the discussions will advance the developments for next-generation methods.

\ifCLASSOPTIONcaptionsoff
  \newpage
\fi

\bibliographystyle{IEEEtranS}
\bibliography{LML}
\begin{IEEEbiography}[{\includegraphics[width=1in,height=1.25in,clip,keepaspectratio]{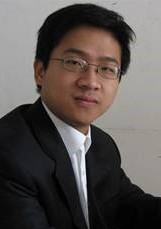}}]{Meng Wang} received the B.E. and Ph.D. degrees in special class for the gifted young from the Department of Electronic Engineering and Information	Science, University of Science and Technology of China (USTC), Hefei, China, in 2003 and 2008, respectively. He is currently a professor with the Hefei University of Technology (HFUT), China. He has authored over 200 book chapters, journal, and conference papers in his research areas. His current research interests include multimedia content analysis, computervision, and pattern recognition. He was a recipient of the ACM SIGMM Rising Star Award in 2014. He is an Associate Editor of the IEEE TKDE, the IEEE TCSVT, and the IEEE TNNLS. \end{IEEEbiography}	
\vskip -2\baselineskip plus -1fil
\begin{IEEEbiography}[{\includegraphics[width=1in,height=1.35in,clip,keepaspectratio]{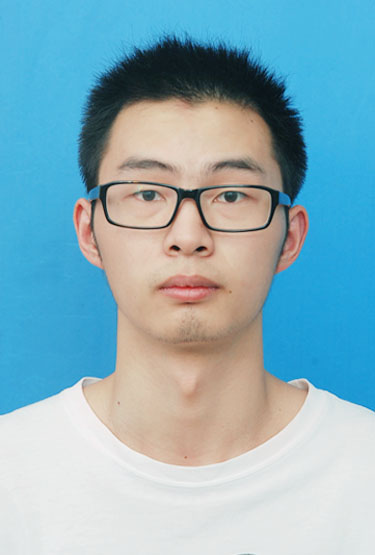}}]{Weijie Fu} is pursuing his Ph.D. degree in the School of Computer Science and Information Engineering, Hefei University of Technology (HFUT). His current research interest focuses on machine learning and data mining. \end{IEEEbiography}
\vskip -2\baselineskip plus -1fil
\begin{IEEEbiography}[{\includegraphics[width=1in,height=1.25in,clip,keepaspectratio]{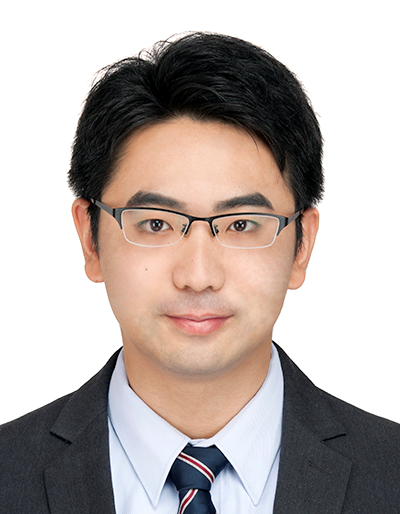}}]{Xiangnan He} is currently a professor with the University of Science and Technology of China (USTC). He received his Ph.D. in Computer Science from National University of Singapore (NUS) in 2016, and did postdoctoral research in NUS until 2018. His research interests span information retrieval, data mining, and multi-media analytics. He has over 50 publications appeared in several top conferences such as SIGIR, WWW, and MM, and journals including TKDE, TOIS, and TMM. His work on recommender systems has received the Best Paper Award Honourable Mention in WWW 2018 and ACM SIGIR 2016. Moreover, he has served as the PC member for several top conferences including SIGIR, WWW, MM, KDD etc., and the regular reviewer for journals including TKDE, TOIS, TMM, TNNLS etc. \end{IEEEbiography}
\vskip -2\baselineskip plus -1fil
\begin{IEEEbiography}[{\includegraphics[width=1in,height=1.25in,clip,keepaspectratio]{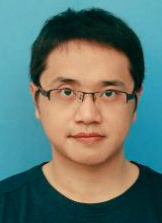}}]{Shijie Hao} is an associate professor at the Hefei University of Technology (HFUT), China. He received his B.E., M.S. and Ph.D. Degree in the School of Computer Science and Information Engineering from HFUT. His current research interests include machine learning and image processing. \end{IEEEbiography}
\vskip -2\baselineskip plus -1fil
\begin{IEEEbiography}[{\includegraphics[width=1in,height=1.25in,clip,keepaspectratio]{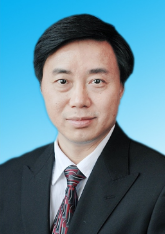}}]{Xindong Wu} is Yangtze River Scholar in the School of Computer Science and Information Engineering at the Hefei University of Technology (HFUT), China, and a Fellow of the IEEE.  He received his Bachelor's and Master's degrees in Computer Science from the Hefei University of Technology, China, and his Ph.D. degree in Artificial Intelligence from the University of Edinburgh, Britain. His research interests include data mining, knowledge-based systems, and Web information exploration. Dr. Wu is the Steering Committee Chair of the IEEE ICDM, the Editor-in-Chief of KAIS, by Springer, and a Series Editor of the Springer Book Series on AI$\&$KP. He was the Editor-in-Chief of the IEEE TKDE between 2005 and 2008. He served as Program Committee Chair/Co-Chair for ICDM '03, KDD-07, and CIKM 2010. \end{IEEEbiography}

\end{document}